%% file: main.tex
\documentclass[sigplan, 10pt]{acmart}
\acmSubmissionID{91}
\renewcommand\footnotetextcopyrightpermission[1]{}
\AtBeginDocument{%
  \providecommand\BibTeX{{%
    \normalfont B\kern-0.5em{\scshape i\kern-0.25em b}\kern-0.8em\TeX}}}

\setcopyright{none}

\acmConference[Conference acronym 'XX]{Make sure to enter the correct
  conference title from your rights confirmation emai}{June 03--05,
  2018}{Woodstock, NY}

\usepackage{subfigure}
\usepackage{subcaption}
\usepackage{multirow}
\usepackage{booktabs}
\usepackage{blindtext}
\usepackage{makecell}
\usepackage{tabularx}
\usepackage{tikz}
\usepackage{amsmath}
\usepackage{amsthm}
\usepackage{algorithm}
\usepackage[noend]{algorithmic}
\usepackage{mathtools}
\usepackage{multirow,multicol}
\usepackage{multicol}
\usepackage{ifthen}

\usepackage{amssymb}
\usepackage{optidef}
\usepackage{wrapfig}
\usepackage{graphicx}
\usepackage{subcaption}
\usepackage{float}
\usepackage{minted}
\usepackage{capt-of}
\usepackage{caption}
\usepackage{enumitem}

\usepackage{cancel}

\newcommand{\sysname}[0]{{Laminar}}

\begin{document}

\title{\sysname: A Scalable Asynchronous RL Post-Training Framework}

\author{
{
Guangming Sheng$^{1*}$, Yuxuan Tong$^{2*}$, Borui Wan$^{1}$, Wang Zhang$^{2}$, Chaobo Jia$^{2}$, Xibin Wu$^{2}$, Yuqi Wu$^{2}$, Xiang Li$^{2}$, Chi Zhang$^{2}$, Yanghua Peng$^{2}$, Haibin Lin$^{2}$, Xin Liu$^{2}$, Chuan Wu$^{1}$ 
}
}
\affiliation{
\vspace{2mm}
{\textit{\Large{$^1$The University of Hong Kong}}\hspace{4mm} \textit{\Large{$^2$ByteDance Seed}}}
\country{}
}

\renewcommand{\shortauthors}{}

\settopmatter{printfolios=true,printacmref=false}

\renewcommand{\shorttitle}{\sysname}
\input{tex/abstract}

\maketitle

\input{tex/introduction}

\input{tex/background}

\input{tex/motivation}

\input{tex/design}

\input{tex/relay}

\input{tex/reorder}

\input{tex/implementation}

\input{tex/experiment}

\input{appendix/appendix_related_work}
\input{tex/conclusion}

\bibliographystyle{ACM-Reference-Format}
\bibliography{reference}

\clearpage
\appendix
\input{appendix/appendix_experiement_detail}

\input{appendix/appendix_experiment_analysis}

\input{appendix/appendix_discussion}

\input{appendix/appendix_comm}

\end{document}

%% file: tex/abstract.tex
\begin{abstract}
Reinforcement learning (RL) post-training for Large Language Models (LLMs) is now scaling to large clusters and running for extended durations 
to enhance model reasoning performance.
However, the scalability of existing RL frameworks is limited,
as extreme long-tail skewness in RL trajectory generation causes severe GPU underutilization. 
Current asynchronous RL systems attempt to mitigate this, but they rely on global weight synchronization between the actor and all rollouts, which creates a rigid model update schedule. 
This global synchronization is ill-suited for the highly skewed and evolving distribution of trajectory generation latency in RL training,
crippling training efficiency.
Our key insight is that efficient scaling requires breaking this lockstep through \textit{trajectory-level asynchrony, which generates and consumes each trajectory independently}.
We propose \sysname, a scalable and robust RL post-training system built on a fully decoupled architecture.
First, we replace global updates with a tier of relay workers acting as a distributed parameter service. 
This enables asynchronous and fine-grained weight synchronization, allowing rollouts to pull the latest weight anytime without stalling the actor's training loop.
Second, a dynamic repack mechanism consolidates long-tail trajectories onto a few dedicated rollouts, maximizing generation throughput.
The fully decoupled design also isolates failures, ensuring robustness for 
long-running jobs. 
Our evaluation on a 1024-GPU cluster shows that \sysname{} achieves up to 5.48$\times$ training throughput speedup over state-of-the-art systems, while reducing model convergence time. 

\end{abstract}

%% file: tex/introduction.tex
\vspace{-1mm}
\section{Introduction} \label{sec:intro}

Reinforcement learning (RL) has emerged as a transformative paradigm for post-training large language models (LLMs), fundamentally enhancing their reasoning capabilities through iterative policy optimization~\cite{deepseek-r1, dapo, post-train_survey}. Contemporary state-of-the-art models, including OpenAI's o1 series~\cite{o1}, DeepSeek-R1~\cite{deepseek-r1}, and xAI's Grok 4~\cite{grok4}, leverage RL techniques to achieve unprecedented performance on complex reasoning tasks spanning mathematics, coding, 
and agentic tasks.

\begin{figure}[t]
    \centering
    \includegraphics[width=\linewidth]{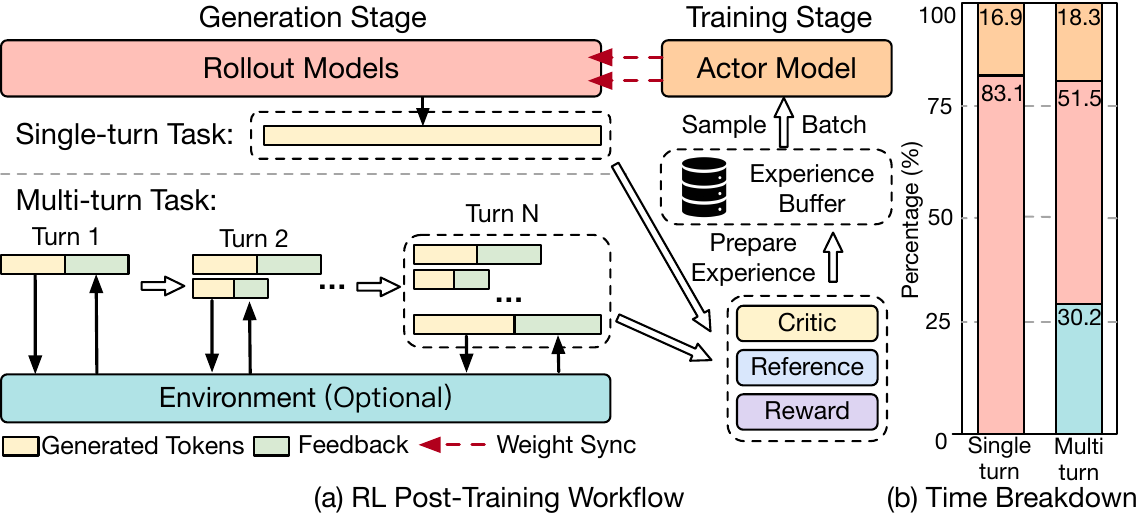}
    \vspace{-8mm}
    \caption{RL post-training workflow and time breakdown of single-turn (math)~\cite{dapo} and multi-turn (code) tasks~\cite{swe-bench}.}
    \label{fig:workflow_breakdown}
    \vspace{-3mm}
\end{figure}

Current RL post-training workflows operate through two main serial phases: generation and training, as shown in Figure~\ref{fig:workflow_breakdown}(a). 
In the generation phase, rollout models produce trajectories by responding to prompts or interacting with environments.
After LLM generates some tokens, the system may trigger an interaction with the environment, which in turn produces feedback.
A recent trend is to scale this phase across many rollout replicas~\cite{deepseek-r1, dapo, seed1.5, k1-5, grok4}. 
xAI scales an RL training job to 
200k GPUs scale
with massive rollout generation~\cite{grok4}.
The goal is to produce vast and diverse trajectories.
These trajectories are then evaluated using verifiers, reward models, or environmental signals~\cite{post-train_survey, reasoning-survey-system1-2} and stored in the experience buffer. 
During the training phase, a subset of these trajectories is sampled from the buffer~\cite{distrl, dapo}.
The sampled batch is then used to compute policy gradients and update model parameters.
This %
extensive generation
is critical for effective policy optimization, 
but relies
on the system's ability to manage the scaled-out rollout process efficiently.

Existing synchronous RL systems~\cite{deepspeed-chat, hybridflow, puzzle, realHF, rlhfuse, openrlhf} face significant efficiency bottlenecks that limit their scalability. The generation stage dominates overall training time, accounting for up to 83.1\% of total execution time in reasoning tasks~\cite{dapo} (Figure~\ref{fig:workflow_breakdown}(b)).
This bottleneck stems primarily from long-tail trajectory generation, where 
the distributions of
output length and environment latency are highly skewed.
For instance, in mathematics and coding tasks, the 99th percentile output length can exceed the 50th percentile by an order of magnitude (Figure~\ref{fig:data_skewness}).
This issue is particularly acute for tasks requiring complex reasoning or multi-turn environment interaction (detailed in \textsection\ref{sec:2_2_background_skewness}).
As generation progresses, only the generation of a few long-tail trajectories remains active, %
leading to severe GPU underutilization across the cluster.
Simply adding more
GPUs cannot resolve this load imbalance issue.

Recent asynchronous RL frameworks attempt to address this long-tail problem by decoupling generation and training across different devices and overlapping their execution~\cite{streamrl, areal, llamarl, rllm2025, intellect-2, asyncflow}. They typically operate under $k$-step bounded staleness, where rollouts use actor weights that are up to $k$ training iterations old.
A global synchronization then distributes the new actor weights to all rollouts following a predefined schedule.
However, this approach still struggles with fundamental inefficiencies on the long-tail generation problem.
A simple and widely adopted one-step staleness ($k$=1) pipeline~\cite{llamarl, streamrl, rllm2025, asyncflow} cannot effectively hide the generation latency of long-tail trajectories.
While increasing the staleness bound ($k$ > 1) can 
absorb this skewness, it 
can negatively affect model convergence~\cite{async-rlhf, dota2, distrl, areal}. 
Consequently, determining an optimal staleness bound that balances overlap efficiency and training convergence remains a difficult tuning problem~\cite{li2018pipe,barkai2019gap, chen2022sapipe, sheng2024mspipe}.

A fundamental limitation of existing RL systems is their reliance on global model weight synchronization, which is not suitable for the 
highly variant and
dynamic nature of RL workloads. 
Recent findings show that trajectory lengths change dynamically as models learn~\cite{deepseek-r1, dapo, k1-5}, exhibiting increasing, decreasing, or fluctuating patterns.
Environment interaction latency varies dramatically due to unpredictable API call time or task complexity~\cite{jimenez2023swe, sandboxfusion, deep-research}.
Current approaches force all trajectories to conform to the same rigid update schedule, regardless of 
their %
diverse characteristics, completion status, or inherent execution variability.
This inflexibility %
enforces rigid data and parameter dependencies across actor and all rollouts,
preventing effective system scaling.

To address these limitations, we propose \sysname, a scalable and robust asynchronous RL post-training system that 
eliminates the long-tail trajectory generation bottleneck at production scale 
while ensuring stable 
RL training.
Our key idea is enabling \textit{trajectory-level asynchrony}, with each trajectory generated and consumed independently at its own optimal pace, to accommodate the vast variability in output length and environment latency.
It removes the primary bottleneck to scaling out to thousands of GPUs and enhances trajectory diversity, which is crucial for efficient RL training.

To realize trajectory-level asynchrony, we introduce a fully decoupled architecture that breaks data and parameter dependencies between the actor model and all rollout replicas, and among rollout replicas themselves.
This architectural decoupling is further fundamental to achieving robustness at scale. By isolating components, the failure of a single rollout machine does not halt the entire training job, enabling swift recovery
that is critical for long-running jobs.

We implement this architecture with two key designs.
{\em First}, we decouple the actor and each rollout replica using a tier of \textit{relay workers}.
Functioning as a distributed parameter service, the relays provide asynchronous and fine-grained weight synchronization. 
The actor can be trained 
uninterruptedly,
while rollouts can retrieve new model parameters from the relays at any time.
Nonetheless, individual rollouts may still face long-tail trajectory 
generation issues.
{\em Further}, we propose a dynamic repack mechanism to consolidate long-tail trajectories from underutilized rollouts 
into a few dedicated rollout replicas, while liberating other rollouts to update using the latest weight version.
This ensures high generation throughput while introducing minimal staleness 
across the system.
Our contributions %
are summarized as follows:

\noindent$\bullet$ We identify trajectory-level asynchrony as the key to scaling-out RL and realize it through a fully decoupled architecture. Our design systematically breaks data and parameter dependencies between all system components and isolates component failures to ensure swift recovery for long-running training (\textsection\ref{sec:overview}).

\noindent$\bullet$ We design a tier of relay workers functioning as distributed parameter service, providing fine-grained, robust, and anytime weight synchronization without stalling training (\textsection\ref{sec:relay}).

\noindent$\bullet$ We propose a dynamic repack mechanism that boosts generation throughput, by actively monitoring rollout idleness via KVCache-based metrics and concentrating long-tail trajectories generation onto fewer rollout replicas
(\textsection\ref{sec:repack}).

\noindent$\bullet$ We conduct extensive experiments comparing \sysname{} with state-of-the-art RL systems~\cite{hybridflow, rllm2025, streamrl, areal} at scales up to 1024 GPUs.
Our evaluation demonstrates up to 5.48$\times$ throughput speedup while ensuring model convergence and robustness
(\textsection\ref{sec:exp}).

%% file: tex/background.tex
\vspace{-2mm}
\section{Background and Motivation} \label{sec:2_bckground}

\subsection{RL for LLM Post-Training} \label{sec:2_1_post_train}
Reinforcement Learning (RL) %
for LLM post-training was first used to
align models with human preferences, known as Reinforcement Learning from Human Feedback~\cite{ouyang2022openai-rlhf, bai2022antrophic-rlhf, daiSafeRLHFSafe2023, zheng2023improving, lee2023rlaif}.
More recently, RL has been extended to enhance complex reasoning in domains like mathematics~\cite{dapo, grpo} and to develop multi-turn agentic capabilities~\cite{swe-bench, retool, deep-research}. 
As shown in Figure~\ref{fig:workflow_breakdown}(a), RL post-training workflow can be primarily decomposed into two stages: Generation and Training.

\noindent\textbf{Generation stage}: 
The rollout model produces trajectories by responding to prompts
through auto-regressive generation or by interacting with environments. 
For single-turn reasoning tasks, such as solving math problems~\cite{AIME}, a trajectory for a simple question may be a short chain-of-thought, while difficult ones can require exploring vast reasoning trees.
For multi-turn agentic tasks like SWE-Bench~\cite{swe-bench}, a trajectory consists of interactions with a code sandbox; fixing a simple bug requires only a few interactions, whereas diagnosing a complex issue can lead to more debugging steps.

\noindent\textbf{Training stage}:
Actor training begins by evaluating each trajectory and assigning it a reward score. 
This score %
can come from a separate reward model~\cite{ouyang2022openai-rlhf, bai2022antrophic-rlhf}, rule-based functions that define explicit criteria~\cite{deepseek-r1, dapo},
or signals derived from the environment~\cite{swe-bench, sandboxfusion}. Each scored trajectory is combined with other metrics, such as advantage estimation 
derived from a reference model and a critic model, to form a training experience. 
These experiences are then added to a buffer, where batches are sampled to update the actor model's parameters via backpropagation. While foundational RL methods like PPO~\cite{ppo} rely on a critic model, many recent algorithms 
(e.g., GRPO~\cite{grpo}, RLOO~\cite{rloo} and DAPO~\cite{dapo})
simplify this process by approximating advantage computation through generating multiple trajectories to the same prompt, removing the need for a critic model entirely.

\subsection{Recent Trends in RL Post-Training}
\label{sec:2_2_background_skewness}

\begin{figure}[t]
    \centering
    \includegraphics[width=\linewidth]{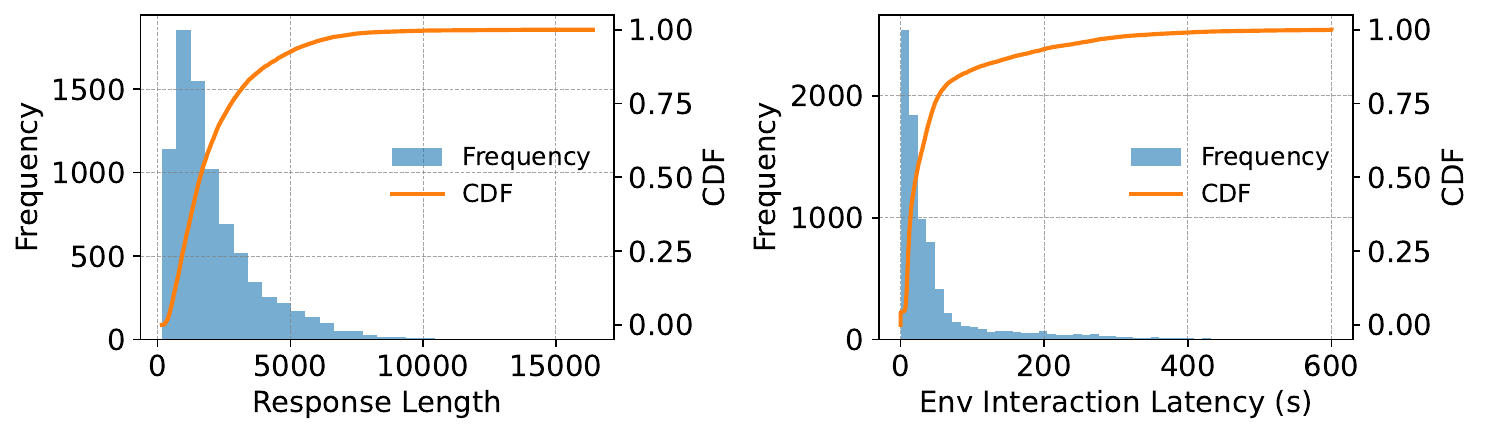}
    \vspace{-7mm}
    \caption{%
    Trajectory length distribution on AIME dataset and execution latency distribution of code sandbox~\cite{AIME, retool}.}
    \label{fig:data_skewness}
    \vspace{-4mm}
\end{figure}

\noindent\textbf{Exacerbated data skewness.}
Data skewness in modern RL post-training has become increasingly severe due to the shift toward reasoning-focused models and agentic tasks.
The primary source of this skew is trajectory length.
As shown in Figure~\ref{fig:data_skewness}, trajectory lengths are highly heterogeneous with the 99th-percentile being ten times the median in some tasks~\cite{dapo, swe-bench}. %
As shorter trajectories in a batch are generated, only a few long-running trajectory generation tasks remain, leading to severe GPU underutilization. The problem is acute because the memory-bound nature of LLM decoding requires large batch sizes to maintain high throughput.
Consequently, the generation stage now dominates the post-training workflow, accounting for up to 83.1\% of total execution time in reasoning tasks, as shown in Figure~\ref{fig:workflow_breakdown}(b).

This long-tail problem extends to multi-turn agentic tasks, which involve highly unpredictable environment execution time.
Here, the LLM interacts with environments like code sandboxes~\cite{swe-bench, retool}, 
computer environments~\cite{mialon2023gaia, xie2024osworld}
and gaming platforms~\cite{dota2, alphastar}.
These environments typically run as external programs or shared services.
As shown in Figure 2, the latency of these interactions can vary significantly due to request queuing and varying computational load. 
This environmental unpredictability, combined with the inherent variance in trajectory length, introduces major efficiency bottlenecks that current systems cannot resolve.

\begin{figure}[t]
    \centering
    \includegraphics[width=\linewidth]{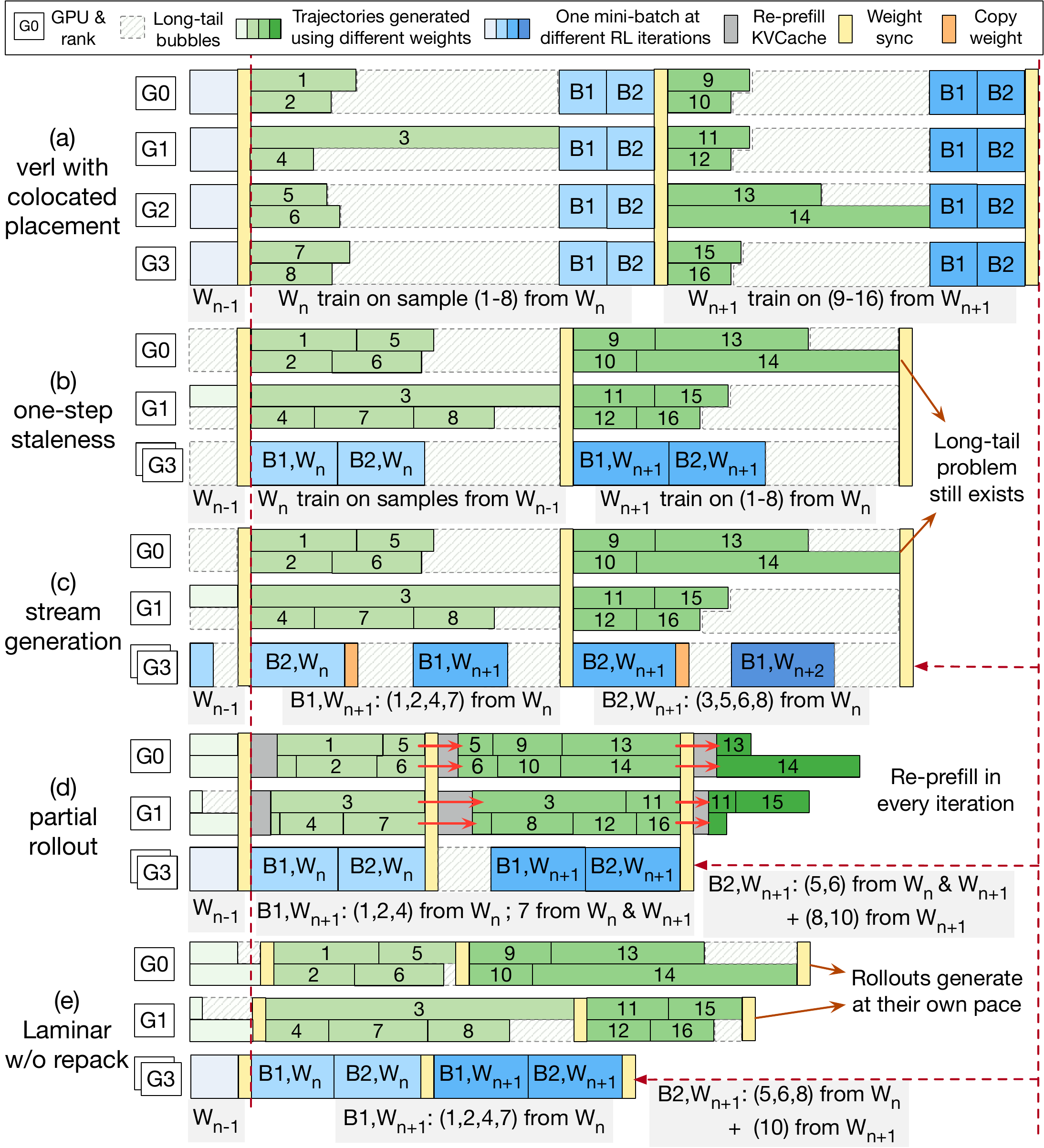}
    \vspace{-7mm}
    \caption{Comparison of RL systems. (a)-(d) adopt GPU-direct communication to perform global weight synchronization, while (c) needs to copy the previous weight version to dedicate memory buffers for later synchronization. (e) performs asynchronous weight synchronization through RDMA on CPU. $\text{W}_\text{n}$ is the actor weight version at iteration n.
    }
    \label{fig:baselines}
    \vspace{-5mm}
\end{figure}

Existing synchronous RL post-training frameworks~\cite{deepspeed-chat, puzzle, hybridflow, realHF, openrlhf, nemo-aligner}
commonly optimize %
the generation stage and the training stage independently.
They adopt sequential and synchronous execution in stages as they are typically built for on-policy RL algorithms.
While some systems adopt hybrid execution parallelism (e.g., HybridEngine~\cite{hybridflow, deepspeed-chat} and Context Switching~\cite{puzzle, realHF}) to improve throughput, they struggle to mitigate the prolonged GPU idle time caused by skewed trajectory generation, as shown in Figure~\ref{fig:baselines}(a).
Other systems~\cite{rlhfuse, adaptive-rlhf} introduce an inter-stage fusion strategy to overlap the generation stage and experience preparation in the training stage.
However, since experience preparation computation only comprises 7.3\% of total RL iteration time,
the benefits remain limited.

\noindent \textbf{Applying asynchronous RL.}
Recent advances in RL post-training have increasingly explored asynchronous (i.e., off-policy) RL algorithms to enable parallel execution of generation and training stages. 
They place actor and rollout models on different devices, allowing trajectories used for training to be generated from previous model versions.
Asynchronous RL algorithms have demonstrated remarkable success in %
domains such as games~\cite{dota2, alphastar} and robotics~\cite{robot, robot2}.
While they improve resource efficiency, their benefits in recent asynchronous RL for LLM post-training systems are severely diminished. 
The fundamental limitation lies in their %
batch-oriented design, which fails to effectively mask the generation latency of long-tail trajectories.

\subsection{Limitations of existing asynchronous RL systems} \label{sec:background_limitation}

\noindent \textbf{Limited scalability due to global weight synchronization.}
Recent asynchronous RL training systems for LLMs pipeline generation and training across different training iterations, ensuring $k$-step staleness bounds%
~\cite{areal, llamarl, streamrl, rllm2025, intellect-2, asyncflow}.
Within an RL iteration, the actor processes a full batch of trajectories by dividing it into smaller mini-batches and conducting a model update on each mini-batch~\cite{ppo, grpo}. %
The actor's final model weights for that iteration are only ready after the last mini-batch is processed.
In a widely used \textit{one-step staleness pipeline}~\cite{rllm2025, streamrl} shown in Figure~\ref{fig:baselines}(b),
rollouts generate a full batch of trajectories using model weights from the past iteration, while concurrently the actor is trained on an older, completely generated batch.
Some concurrent research~\cite{streamrl, asyncflow} introduces a \textit{streaming generation} approach.
As shown in Figure~\ref{fig:baselines}(c),
the actor is trained first on short trajectories in early 
mini-batches 
while consuming long trajectories in later 
mini-batches.
Both systems create a strong data dependency: trainer's progress is tied to the completion of a full batch, forcing it to wait for the slowest trajectory and coupling the execution of all rollouts.

These systems also suffer from a rigid model weight dependency, as all rollouts rely on a global synchronization point to receive new weights after $k$ training iterations.
This global synchronization becomes a major bottleneck in tasks with highly skewed generation times
(e.g., math competition~\cite{AIME}), 
failing to effectively hide the long generation time of a few tail trajectories and creating substantial pipeline bubbles.
It limits
scalability, as adding more GPUs to increase rollouts cannot %
mitigate the wait for long-tail generation,
and allocating additional GPUs per rollout 
provides only marginal latency reductions,
as shown in Figure~\ref{fig:latency_bsz}.

Other asynchronous RL systems~\cite{seed1.5, areal, k1-5, llamarl} improve the global synchronization design with \textit{partial rollout}. 
It interrupts all ongoing trajectories generation to apply the latest actor model in rollouts, and then continues trajectory generation using the latest model weights (Figure~\ref{fig:baselines}(d)). 
A single long trajectory is then composed of several segments, each generated by a different policy version.
While this reduces long-tail bubbles, it introduces two main issues. (1) The pause-and-sync cycle incurs significant overhead by forcing rollouts to rebuild the KVCache (i.e., re-prefill) for every interrupted trajectory, repeatedly in each RL iteration,
wasting GPU resources without advancing generation.
(2) Generating a single response with inconsistent policy versions
can harm model convergence,
resulting in slower convergence as we show experimentally in \S\ref{sec:exp_train_convergence}.
A comprehensive discussion with more related works is provided in \S\ref{appendix:related_work}.

\noindent \textbf{Undesirable coupling between system throughput and training stability.} 
A core challenge in existing asynchronous RL systems is selecting an appropriate staleness bound $k$.
A larger $k$ can better hide generation latency and improve system throughput. However, it also increases the divergence between the data-generating policy and the policy being trained, which can harm model convergence.
Conversely, a smaller $k$ reduces this divergence but is less effective at masking generation latency, leading to poor throughput.
Finding a suitable $k$ is %
a difficult, task-specific tuning problem.

This %
problem is compounded by the dynamic nature of RL training. As an LLM learns, trajectory length often changes significantly~\cite{deepseek-r1, k1-5, dapo}. %
A staleness bound $k$ that is optimal early in training can become inefficient or unstable over time. Existing systems treat $k$ as a static hyperparameter, 
creating a rigid parameter dependency
that fails to adapt to evolving training dynamics.

%% file: tex/motivation.tex
\vspace{-3mm}
\subsection{Opportunity and Challenges} \label{sec:3_motivation}

\noindent\textbf{Opportunity: Adopting trajectory-level asynchrony to scale up RL.}
Existing asynchronous RL frameworks are fundamentally constrained by a global synchronization point to update all rollouts. 
The key opportunity lies in decoupling individual trajectories generation from this global lockstep, %
enabling \textit{trajectory-level asynchrony}. 
It allows each trajectory generation to complete at its own pace on a consistent rollout model version and be stored in an experience buffer.
Rollouts operate independently and do not stall the execution of one another
(Figure~\ref{fig:baselines}(e)).
This effectively masks generation delays of longer trajectories and
allows shorter trajectories to be promptly available for sampling.
The fully decoupled trainer samples a batch from the experience buffer
without interrupting ongoing rollout generation.
Importantly, the staleness of each trajectory (due to the model version it is generated on) emerges naturally from its %
generation latency, rather than due to a %
static staleness bound (%
\textsection\ref{sec:asynchrony_analysis}).
The staleness %
varies according to evolving trajectory lengths and diverse environmental latencies throughout training.

We identify the following challenges in achieving this trajectory-level asynchrony and scaling up RL.

\vspace{1mm}
\noindent\textbf{Challenge 1: Support asynchronous weight synchronization between actor and %
rollouts.}
Implementing tra- jectory-level asynchrony requires %
asynchronous and fine-grained weight synchronization,
where a rollout fetches the latest actor weights as soon as it completes a batch's generation.
As each rollout operates independently and finishes generation at its own pace, %
rollout updates can occur at any moment during actor training.
Existing asynchronous RL systems leverage high GPU-GPU bandwidth for parameter transfers 
at global synchronization points~\cite{streamrl, llamarl, areal}.
This approach becomes infeasible for asynchronous rollout updates.

\textit{First}, direct GPU transfers for %
asynchronous rollout updates introduce resource contention. 
It requires dedicated memory buffers, yet GPU memory is a scarce and critical resource in RL training~\cite{hybridflow, realHF, rlhfuse}. To maximize throughput, both actor and rollouts operate near peak memory capacity; actor uses large training batches, while rollouts expand their KVCache for larger decode batches~\cite{mooncake}. However, asynchronous transfers require buffering. As discussed in \textsection\ref{sec:background_limitation}, the final model weights from an RL iteration are not available until the last mini-batch has been trained. Consequently, to support on-demand pulls from rollouts during a training %
iteration, the actor must hold the previous iteration's weights in a buffer. Similarly, to handle proactive pushes from the actor, a rollout must buffer incoming weights until its batch generation completes.
For large models, the buffer can exceed available GPU memory. Even if it fits, this would reduce the training batch size or KVCache capacity, degrading system throughput.
Furthermore, the communication kernels (e.g., NCCL) 
may compete for the GPU’s streaming multiprocessors with training and decoding kernels, stalling computation on both actor and rollouts.

\textit{Second}, without these buffers, the actor must stall until rollouts fetch the new weights 
before it can proceed to the next mini-batch's training, in %
an RL iteration.
This reintroduces a blocking synchronization point, defeating the purpose of trajectory-level asynchrony, especially as the number of rollouts increases.

We seek to minimize GPU idle time 
while
introducing zero extra GPU memory consumption
on both the actor and rollouts during %
asynchronous weight synchronization, %
by introducing a tier of relay workers
(\textsection\ref{sec:relay}).

\begin{figure}[t]
    \centering
    \includegraphics[width=\linewidth]{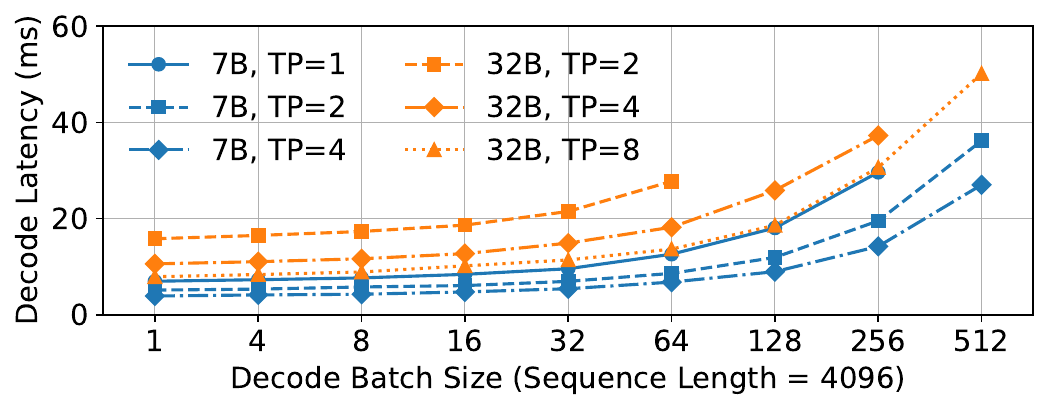}
    \vspace{-7mm}
    \caption{
    One-step decode latency of Qwen2.5-7B/32B on H800 GPUs, under various tensor parallel (TP) sizes with decode batch sizes %
    up to the KVCache limit.
    }
    \label{fig:latency_bsz}
    \vspace{-4mm}
\end{figure}

\vspace{1mm}
\noindent\textbf{Challenge 2: Resolve the long-tail generation problem at each rollout.}
While the relay worker design allows rollouts to operate independently, individual rollouts can still get stuck in long-tail generation
(Figure~\ref{fig:baselines}(e)).
Some rollouts may eventually run only a few %
trajectories' generation, 
resulting in low utilization of their GPUs.
More critically, these rollouts are prevented from fetching the latest policy (model weights) from the actor, forcing them to continue generating trajectories with increasingly stale weights. 
Maximizing the number of rollouts that generate near-on-policy trajectories is crucial for training stability and model performance.

We leverage %
that LLM decoding is a memory-bound operation~\cite{vllm, nanoflow, sglang}.
Rollout replicas stuck on long-tail generations are left with a very small decode batch (e.g., <8) of incomplete trajectories. For a memory-bound operation,
decoding such a small batch has nearly the same latency as a much larger one (e.g., 64)~\cite{Sarathiserve, llumnix, megascale-infer},
as shown in Figure~\ref{fig:latency_bsz}.
This motivates us to consolidate incomplete trajectories from multiple rollouts using the same weight version onto a few destination rollouts, forming a larger decode batch with negligible impact on latency.
The released replicas can then be updated with the latest actor weights to produce fresh on-policy trajectories.

However, the dynamic nature of RL training complicates this consolidation. 
The central challenge is determining when to trigger consolidation and which rollout replicas are involved,
as identifying genuinely underutilized rollouts is non-trivial.
Prior research has relied on task- and model-specific metrics that require extensive offline profiling to define static thresholds to identify underutilization~\cite{rlhfuse}, which are impractical for dynamic RL workloads
due to the prohibitive overhead of repeated profiling.
We design an efficient solution that uses a lightweight indicator based on KVCache utilization to dynamically make consolidation decisions
(\textsection\ref{sec:repack}).

\vspace{1mm}
\noindent\textbf{Challenge 3: Ensure robust RL training at scale.}
Production-level RL post-training jobs are scaling to thousands of GPUs and can run for weeks or months~\cite{grok4, k2}. 
At this scale, hardware faults are inevitable. 
As %
the rollout generation stage often dominates RL iteration time,
resilience of rollouts is critical to overall training progress.
While the trainer can leverage checkpointing for recovery, existing RL systems provide no fault tolerance for the rollout stage.
In addition, %
GPU-direct communication with NCCL~\cite{nccl} is typically used for weight synchronization~\cite{areal, streamrl, openrlhf, hybridflow}, and NCCL lacks native support for fault tolerance and elasticity~\cite{demystifying-nccl}.
Consequently, a single rollout failure can force a full job to restart from a checkpoint, preventing fast recovery needed for long-running workloads.

In existing systems, each rollout independently manages its active trajectories being generated. A machine failure results in the complete loss of all in-progress work on that node. 
This loss is particularly costly for complex tasks~\cite{swe-bench, deep-research}, where a single trajectory can take hours to generate.
Regenerating trajectories during failure recovery wastes valuable GPU cycles. %
Our fully decoupled architecture enables robust long-running RL training by isolating each system component, ensuring rapid and independent recovery
(\textsection\ref{sec:4_3_fault_recovery}, \textsection\ref{sec:relay_fault_tolerant}).

%% file: tex/design.tex
\section{Fully Decoupled Architecture Design} \label{sec:overview}

We introduce \sysname, a scalable and robust asynchronous RL post-training framework designed to scale out RL training jobs, while ensuring robust %
training.
\sysname{} achieves full decoupling of both data dependencies (i.e., decoupling trainer consumption from %
trajectory batch generation) and parameter dependencies (i.e., no lock-step weight synchronization)
between the actor 
and all rollout replicas, as well as among rollouts themselves.
Such decoupling enables trajectory generation, actor model training, and weight synchronization between actor and rollouts to proceed independently.
It eliminates any global synchronization bottlenecks that limit scalability, and effectively isolates faults to ensure rapid, non-disruptive recovery and long-running training.
Enabling trajectory-level asynchrony, our framework allows each trajectory's generation to complete at its own optimal pace, while maintaining training stability with diverse trajectories under minimal staleness.

\subsection{System Components}
\begin{figure}[t]
    \centering
    \includegraphics[width=\linewidth]{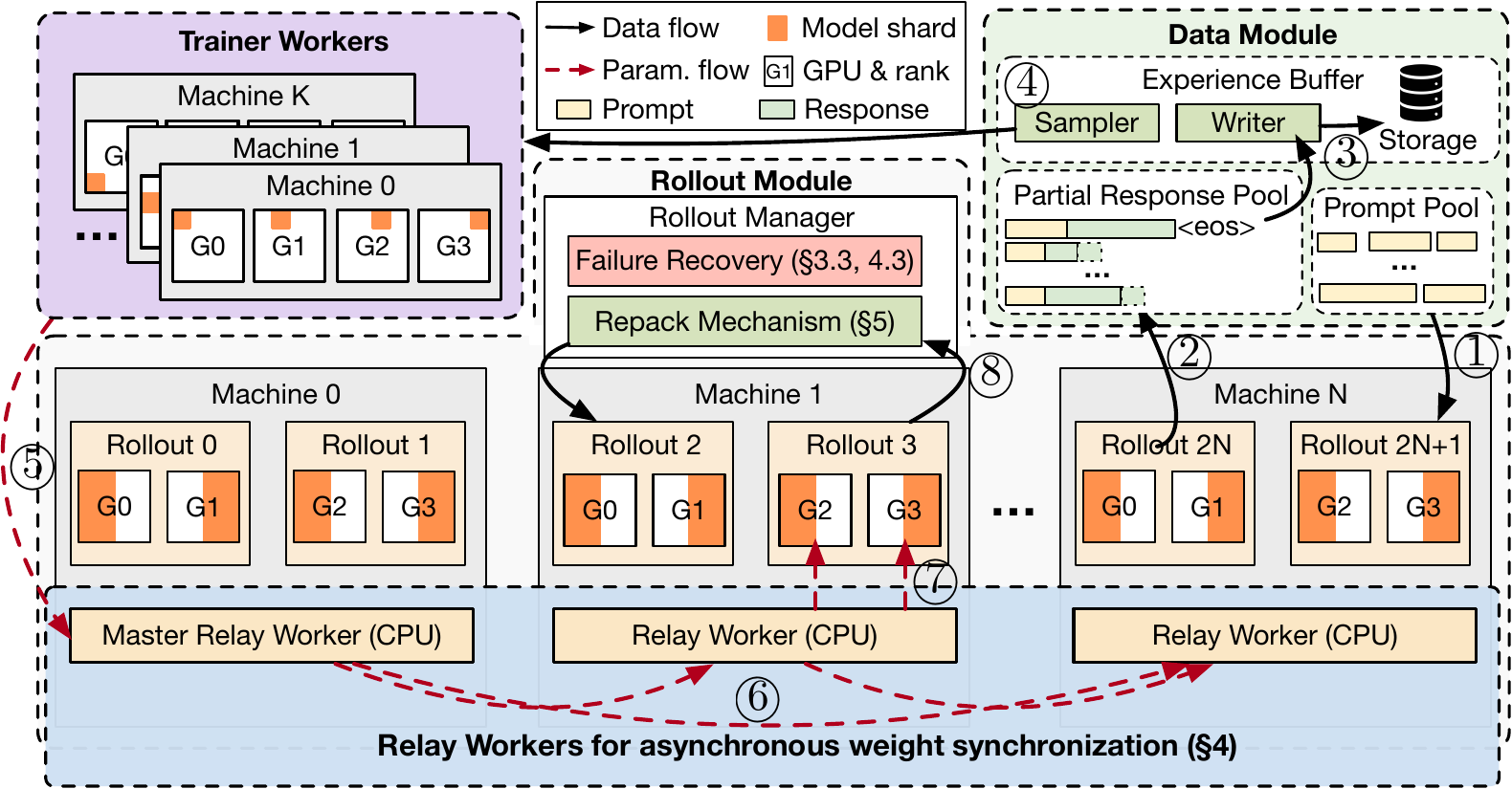}
    \vspace{-7mm}
    \caption{\sysname{} %
    architecture and training workflow.
    }
    \label{fig:workflow_arch}
    \vspace{-3mm}
\end{figure}

Figure~\ref{fig:workflow_arch} gives the architecture of \sysname{}, consisting of four core modules:

\noindent $\bullet$\textit{Rollout Module} consists of a rollout manager and numerous rollouts.
The manager runs on a CPU machine, isolating it from any failures of GPU machines.
It coordinates the rollouts by monitoring their workload and applying trajectory repacking accordingly.
It also ensures system stability by monitoring rollout health, managing the weight update topology, and handling fault recovery.
Each rollout performs auto-regressive generation to produce trajectories independently and may interact with external environments during generation. 
After generating its own batch sampled from the prompt pool,
each rollout fetches the latest actor model weights from its colocated relay worker.

\noindent $\bullet$\textit{Data Module} manages the lifecycle of trajectories through three distinct storage components, each running as a separate process and storing data in a CPU machine: 
a prompt pool to supply initial states for generation, such as a math or coding question; 
a partial response pool that centrally stores 
in-progress trajectories to ensure fault tolerance; and an experience buffer that holds completely generated trajectories. 
Interaction with this buffer is managed by a writer and a sampler.
Flexible APIs are provided for both writer and sampler,
allowing users to customize the sampling strategy for training and the eviction strategy for removing old experiences when buffer capacity is not enough.

\noindent $\bullet$\textit{Relay Workers} function as a hierarchical parameter service.
The rollout manager designates one relay as the master, which receives updated weights from the actor and broadcasts them to the other relays.
Each relay is a separate CPU process running on each rollout GPU machine, hosting the latest actor model weights in local CPU memory. %
This allows any rollout replica to pull a new version on demand without blocking GPU computation on trainer and other rollouts. 

\noindent $\bullet$\textit{Trainer} %
samples batches of experiences from the experience buffer to perform model updates according to the selected RL algorithm, and may involve multiple models, such as actor, critic, and reference models~\cite{ppo, grpo}.
The models are colocated on the same set of GPUs, and are executed sequentially in a time-sharing manner~\cite{hybridflow, deepspeed-chat}.

\subsection{Training Workflow}
The continuous, asynchronous training workflow of \sysname{} is %
designed to maintain high training throughput when scaling up.
It begins as rollouts pull prompts from the prompt pool to generate trajectories on their GPUs
(step \textcircled{1}). 
For fault tolerance, in-progress trajectories are streamed to the partial response pool (step \textcircled{2}). Upon generation completion, they are moved to the experience buffer (step \textcircled{3}).
In parallel with rollout generation, the trainer %
samples completed trajectories from the experience buffer 
to perform model training (step \textcircled{4}). %
This fundamental decoupling of data production from consumption is key to the system's scalability. 

After a model update, the trainer pushes the new actor weights to the master relay and immediately resumes its next training iteration without waiting for weights to be fully distributed to other relays or rollouts (step \textcircled{5}). 
The master relay then broadcasts the new weights directly to all other relays using RDMA, which occurs in the background on CPU memory without affecting ongoing GPU-based generation on the same machine (step \textcircled{6}). 
A rollout can fetch the latest weights from its colocated relay at any time, over high-speed PCIe with minimal latency (step \textcircled{7}). 
This decouples parameter dependencies across the actor and all rollouts in the system.

Meanwhile, \sysname{} actively manages workload imbalance on rollouts to maintain high generation throughput. Rollout manager monitors rollouts stuck on long-tail trajectories generation and triggers a repack mechanism (step \textcircled{8}). 
The repack consolidates trajectories from underutilized rollouts onto fewer rollout, freeing the former which can pull the latest model weights and reducing system-wide staleness.

\vspace{-2mm}
\subsection{Fault Tolerance and Recovery} \label{sec:4_3_fault_recovery}

The decoupled architecture in \sysname{} 
allows dynamic removal and addition of rollouts without halting training and losing data, attaining elasticity. %
Individual faults in rollouts, relays, and the trainer are isolated,
enabling rapid recovery without a costly global restart.
This resilience is critical for large-scale training and is handled in key stages of the training workflow.

When a fault occurs during rollout generation, 
our heartbeat-based failover mechanism 
quickly detects a faulty rollout and notifies the rollout manager. %
We first attempt to recover by re-initializing the faulty replica on the same GPUs. If the fault persists after re-initialization, we then evict the entire affected machine. During this, in-progress trajectory states remain safe in the partial response pool.
The rollout manager redirects interrupted trajectories to healthy rollouts with the same weight version, or waits for replacement machines if none are available. 
Recovery is swift as new machines initialize rollouts and corresponding relays by synchronizing with the master relay for the latest weights 
or loading specific weight versions from actor checkpointing files. %

Relay faults during weight synchronization are managed by our fault-tolerant relay broadcast mechanism (\textsection\ref{sec:relay_fault_tolerant}). Faulty relay workers are detected instantly without a timeout, and the communication scheduler rebuilds the broadcast chain using healthy nodes. 
This recovery can be completed within seconds without disrupting ongoing rollout generation.

Trainer faults are handled by standard checkpoint recovery methods~\cite{wagenlander2024tenplex, bytecheckpoint, phoenixos}, with actor model weights checkpointed periodically. When a trainer worker fails, it is evicted %
~\cite{amazon, characterization} and then recovered from the latest checkpoint. During this period, rollouts continue generation with the latest available weights.
Once recovered, the actor resumes sampling from the experience buffer and resumes training.

%% file: tex/relay.tex
\section{%
Asynchronous Weight Synchronization Using Relay Workers} \label{sec:relay}

\begin{figure}[t]
    \centering
    \includegraphics[width=\linewidth]{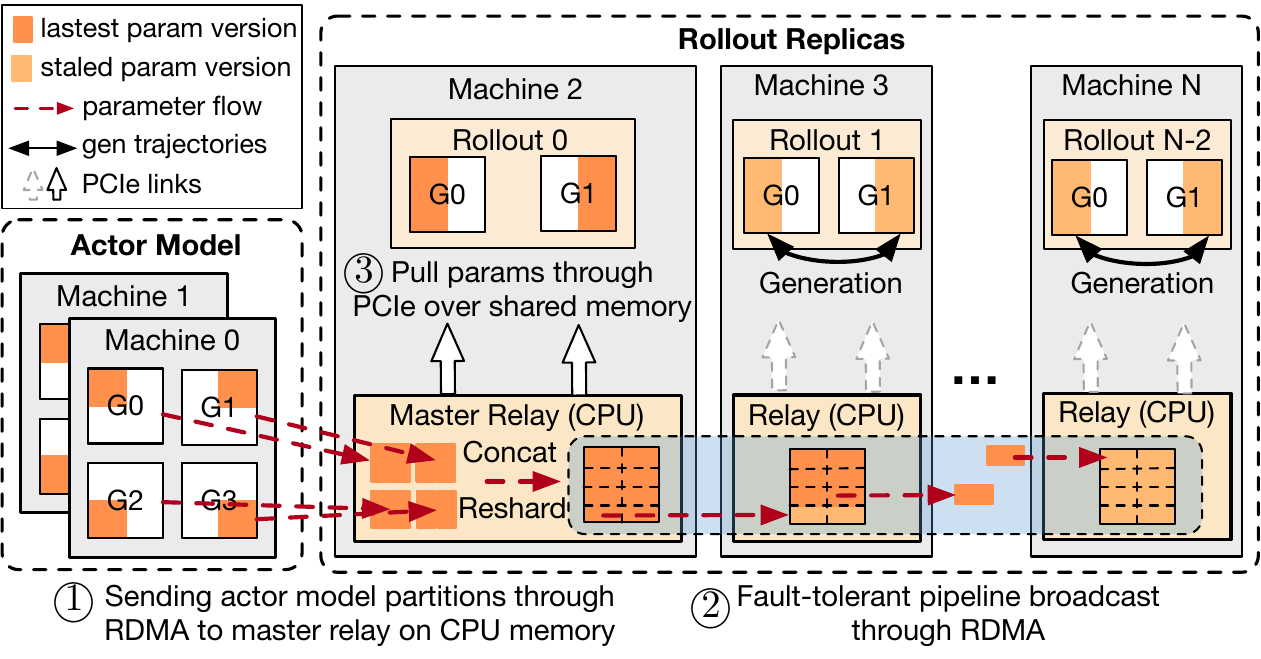}
    \vspace{-7mm}
    \caption{%
    Asynchronous weight synchronization workflow.}
    \label{fig:relay_workflow}
    \vspace{-3mm}
\end{figure}

\subsection{Design Considerations} \label{sec:relay_design_consideration}
\noindent \textbf{%
Inefficiency of using a storage system.}
Traditional RL systems often involve a storage system for weights synchronization between the actor and rollout models. 
SRL~\cite{srl} relies on a file system (%
NFS~\cite{nfs}) and OpenAI-Five~\cite{openaifive} utilizes key-value stores (%
Redis~\cite{redis}) to transfer and store the published weights.
When model sizes were measured in megabytes, weight synchronization through these storage systems incurs reasonable delay; however, they are ill-suited for LLMs with gigabytes scale of model sizes. 
{\em First}, substantial serialization and I/O overhead is incurred.
Our profiling of a 32B LLM shows that serializing a single 4GB model shard takes approximately 8 seconds.
Transferring this volume of data over a standard %
TCP network to/from the storage system
adds another 10 to 20 seconds of latency into the critical path of every rollout.
{\em Next}, the storage system renders a contention bottleneck when numerous rollouts simultaneously pull the latest
weights, %
incurring long and unpredictable latencies for weight updates.
Instead, we exploit local host memory and high-bandwidth networks like RDMA for efficient weight synchronization, avoiding serialization and a single point of contention.

\noindent \textbf{%
Hosting relay workers in local host memory.}
Direct GPU-to-GPU transfers for %
asynchronous weight synchronization %
incur prohibitive memory overhead and computational stalls (\textsection\ref{sec:3_motivation}).
Allocating separate GPUs dedicated as relays would be computationally wasteful and cost-prohibitive.
We introduce a tier of relay workers that run in CPU memory of rollout machines,
exploiting %
that such host memory remains largely underutilized.
The CPU-based intermediaries decouple the actor model weights from asynchronous rollout demands and leverage RDMA for efficient and robust weight synchronization among machines.

\subsection{Hierarchical Relays and Their Workflow} \label{sec:relay_broadcast}

Our relay worker hierarchy is designed to attain two goals: \textbf{1)} to minimize actor stall time during weight publication, and \textbf{2)} to ensure low-latency access for any rollout requesting a new version. 
For the first goal, we designate a single master relay: the actor transfers its updated weights to this master relay and can immediately resume training, effectively hiding the broadcast latency (step $\textcircled{1}$ in Figure~\ref{fig:relay_workflow}).
After receiving the latest weights, the master relay reshards them according to the rollout sharding strategy (e.g., tensor parallelism among GPUs a rollout runs on)
~\cite{hybridflow, deepspeed-chat, puzzle, realHF}.
A single master relay may become a bottleneck when serving weight requests from multiple rollouts. We further adopt one relay worker per rollout machine to distribute the workload, and enable rollout's low-latency access to updated weights anytime.

The updated weights must be efficiently propagated from the master relay to all other relays. We implement this with
a chain-based pipelined broadcast 
using RDMA (step $\textcircled{2}$).
The broadcast pipelines model chunk transfers along a chain of relays, overlapping communication among different hops. %
This makes the broadcast time nearly constant regardless of the length of the chain scale~\cite{chain-broadcast}, which we formally analyze in Appendix~\ref{sec:appendix_broadcast}.
Broadcast time of less than 1.6 seconds is incurred for a 72B model from the master to 127 
other relays %
(Figure~\ref{fig:relay_broadcast_latency} in Appendix~\ref{sec:appendix_broadcast}).
This broadcast time is negligible compared to the lengthy trajectory generation time at hundreds to thousands of seconds~\cite{dapo, retool, swe-bench}.
Rollouts fetch the latest weights through PCIe links from their colocated relay at any time, without waiting for the resharding and broadcast to complete (step $\textcircled{3}$).

\begin{figure}[t]
    \centering
    \includegraphics[width=\linewidth]{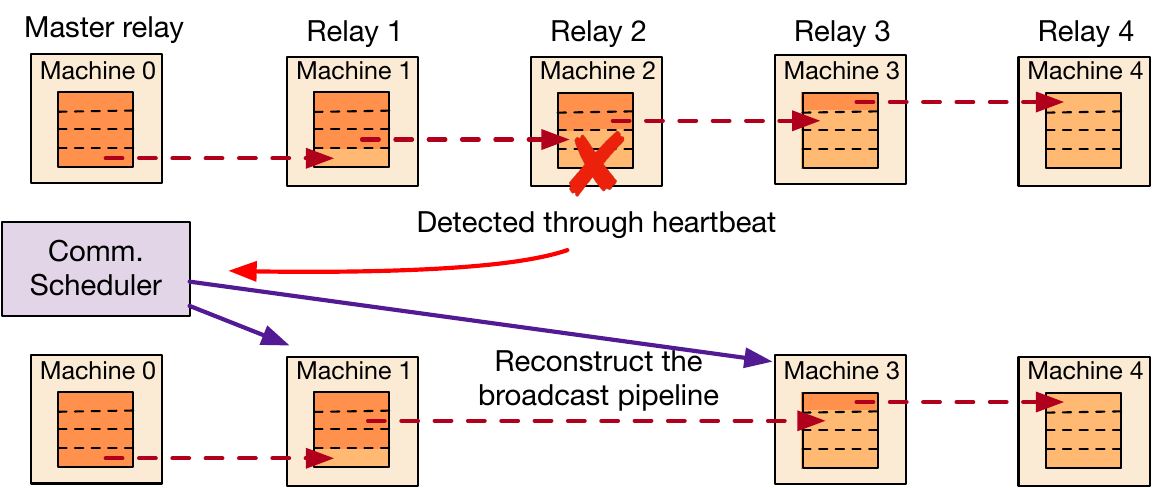}
    \vspace{-6mm}
    \caption{Swift recovery during relay broadcast.}
    \label{fig:fault_relay}
    \vspace{-3mm}
\end{figure}

\subsection{Fault-Tolerant Relay Broadcast} \label{sec:relay_fault_tolerant}

As RL training scales to thousands of GPUs, hardware and software faults become inevitable. 
We design a fault-tolerant relay broadcast scheme. %
The communication scheduler in
the rollout manager
establishes the broadcast pipeline connecting the relay workers 
on
rollout machines. 
The master relay receives weights from the actor, chunks them, and broadcasts them down the chain in a pipelined manner.
If a rollout machine fails, the scheduler detects the failure via heartbeat monitoring.
The failed machine is evicted and the scheduler immediately reconstructs the broadcast chain 
among the remaining rollout machines, as depicted in Figure~\ref{fig:fault_relay}.
This repair is a constant-time ($O$(1)) operation that can be completed in less than one second. 
If the master relay fails, the rollout manager selects a new master from the available relays and rebuilds the broadcast chain. The trainer is then notified of the new master relay's IP address, for resuming weight distribution.
During this fast repair, RL training continues seamlessly, %
and rollout generation on healthy rollout machines is not affected, as rollout generation and relay communication are isolated on different processes in each machine (\textsection\ref{sec:impl}).

%% file: tex/reorder.tex
\section{Bubble Elimination in Long-tail Trajectory Generation%
} \label{sec:repack}

Each rollout fetches the last model weight upon its completion of a trajectory batch generation. At any time, %
different versions of the model weights can be used by different rollouts in trajectory generation, while groups of rollouts are performing generation using the same weight versions. 
To remove GPU idle time on rollouts generating long-tail trajectories, we introduce a repack mechanism that consolidates in-progress trajectory generation from these straggler rollouts onto a few designated rollouts within the same weight version group.
Our design involves three key aspects: establishing the workflow of repacking, monitoring rollouts trapped in long-tail generation, and determining suitable repack destination rollouts.

\begin{figure}[t]
    \centering
    \includegraphics[width=\linewidth]{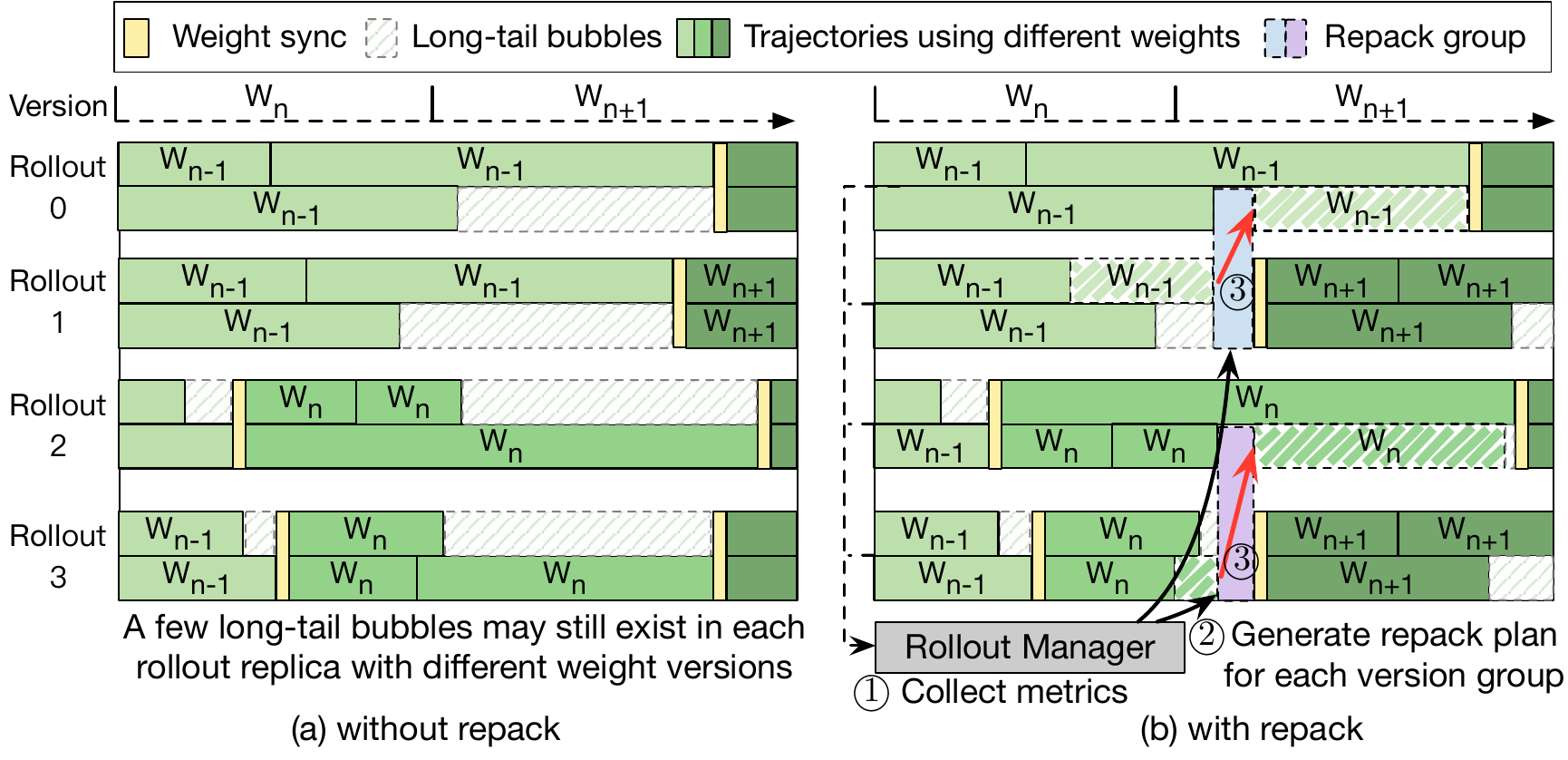}
    \vspace{-7mm}
    \caption{Workflow of the Repack Mechanism. Dashed timeline shows the weight version in the trainer.
    }
    \label{fig:repack_workflow}
    \vspace{-3mm}
\end{figure}

\vspace{-1mm}
\subsection{Workflow of the Repack Mechanism} \label{sec:6.1_repack_workflow}

Our repack mechanism is primarily triggered by a periodic check %
(e.g., 5 seconds),
which detects if any rollout replicas are trapped in long-tail trajectories generation and thus being underutilized. %
Additionally, %
a repack is also initiated immediately after the trainer completes weight updating on a global batch,
aiming to more rapidly free up rollouts for %
on-policy trajectories generation with the latest model version.

The repack workflow, given in Figure~\ref{fig:repack_workflow}, begins with the rollout manager collecting progress metrics from all rollouts 
and grouping them by their model weight versions
(step $\textcircled{1}$).
Within each group, the rollout manager 
identifies rollouts undergoing long-tail generation
using an idleness metric (\textsection\ref{sec:6.2}). %
Then a %
packing algorithm is performed to decide the repacking plan that consolidates in-progress trajectories from multiple rollouts into fewer destination rollouts (step $\textcircled{2}$). %
The rollout manager %
then transfers unfinished trajectories of the long-tail generation rollouts to destination rollouts accordingly (step $\textcircled{3}$).

\begin{figure}[t]
    \centering
    \includegraphics[width=\linewidth]{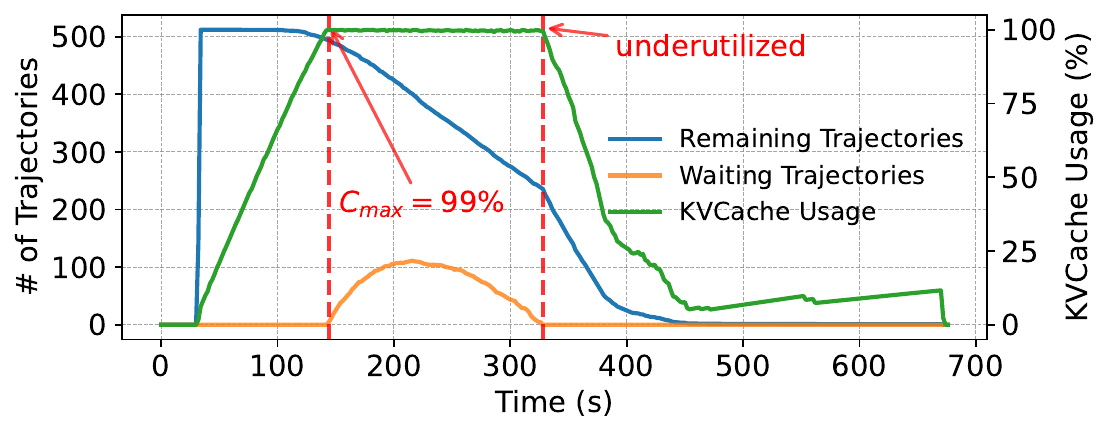}
    \vspace{-7mm}
    \caption{KVCache utilization lifecycle during rollout generation. Usage ramps up to a threshold, then remains steady while remaining trajectories decrease, and finally falls, marking the idle phase that enables trajectory repacking.}
    \label{fig:kvcache_usage}
    \vspace{-3mm}
\end{figure}

\subsection{Online Trajecgtory Repacking} \label{sec:6.2}

Upon each triggering, the %
packing algorithm dynamically partitions a group of rollouts into sources (conducting long-tail generation to be released) and destinations (to consolidate long-tail trajectories generation onto). %

\vspace{1mm}
\noindent \textbf{Idleness metric.}
We design the idleness metric to identify GPU-underutilized rollouts.
Prior approaches rely on a predefined threshold of remaining generation requests to detect the long-tail effects~\cite{rlhfuse}. 
Determining an optimal threshold requires extensive, per-job offline profiling, which is impractical for RL deployments where workloads can vary significantly.
Instead, we leverage \textit{KVCache utilization} as a more direct indicator of resource pressure in memory-bound LLM generation.
Our observation is that KVCache capacity is the primary barrier to scaling the parallel-decode batch size in memory-bound rollout generation~\cite{vllm, nanoflow, sglang}.
We have observed that the latency per decode step remains stable even as the batch size increases (Sec.~\ref{sec:3_motivation}).

Figure~\ref{fig:kvcache_usage},
whose results are from a 32B model generating a batch of 512 trajectories on H800 GPUs (TP=4),
shows that a rollout's KVCache usage follows a distinct lifecycle according to a natural threshold, $C_\text{max}$ (e.g., above 99\% of total KVCache),
which indicates full KVCache utilization.
Initially, usage ramps up to this threshold as the rollout is filled with active token generation of trajectories.
KVCache utilization then remains at this peak even as trajectories finish. This is because waiting trajectories in the batch immediately fill the newly available KVCache space. The KVCache usage begins to fall from this peak only when no waiting trajectories are left and the remaining running trajectories complete.
We consider the rollout idle from this time on and its remaining long trajectory generations become candidates for repacking. 
This consistent threshold behavior across different RL workloads eliminates the need for workload-specific threshold tuning, facilitating our identification of source rollouts. %

\noindent \textbf{%
Repacking algorithm.} 
We consider trajectory repacking as a %
bin packing problem~\cite{fast-bin-packing, new-bin-packing}, where each underutilized rollout represents an item 
and each destination rollout acts as a bin.
The destinations are selected from the pool of underutilized rollouts
to achieve 
the most densely packed KVCache after the repack.
Our primary goal is to maximize the number of released source rollouts (aka minimize the number of destination rollouts) 
with minimal overhead and negligible impact on generation latency (verified in \textsection\ref{sec:exp_repack}).
We design an efficient %
heuristic algorithm (Algorithm~\ref{alg:repack}), inspired by the Best-Fit approach~\cite{worst-bin-packing-best-fit}. 

We define a rollout's capacity using two key indicators: its KVCache utilization, $C_\text{used}$, and its current trajectory count $N_\text{reqs}$. 
The KVCache threshold, $C_\text{max}$,
serves as a memory capacity limit that bounds the decode batch size, representing the maximum number of trajectories a single decode step can hold.
We also enforce a batch size upperbound, $B$, 
representing the maximum number of trajectories that can be decoded in parallel with only a negligible increase in latency.
This limit is determined using the roofline model~\cite{williams2009roofline, llm-roofline}, which identifies the point where the decode operation transitions from being memory-bound to compute-bound. Exceeding this limit would unacceptably increase generation latency. 
A rollout becomes one of the candidates $S$
for repacking if it is in its ramp-down phase (its KVCache utilization is non-increasing and below the $C_\text{max}$ threshold)
and its remaining trajectory count is smaller than the upperbound $B$ (Line 3).
Then we sort the candidate rollouts $S$ and prioritize releasing those with the smallest KVCache footprint first (Line~\ref{line:sort}), as smaller workloads are easier to be replaced.

\begin{algorithm}[t]
\small
\caption{Best-Fit Trajectory Consolidation}
\label{alg:repack}
\begin{algorithmic}[1]
\STATE {\bfseries Input:} Set of rollout replicas $R$, KVCache threshold $C_\text{max}$, Roofline batch size $B$
\STATE {\bfseries Output:} A consolidation plan $P$ of (source, destination) pairs
\STATE $S \leftarrow \{r \in R \mid r.C_\text{used} < \text{min}(C_\text{max}, r.C_\text{prev}) \land r.N_\text{reqs} < B\}$ \label{line:candidate}
\STATE $S \leftarrow \text{sort}(S, \text{key} \leftarrow r.C_\text{used})$ \label{line:sort}
\STATE $P \leftarrow \emptyset$; $\mathcal{E} \leftarrow \emptyset$ \textit{ // Plan and set of emptied replicas}
\FORALL{$s \in S$} \label{line:start_for}
    \IF{$s \in \mathcal{E}$} 
        \STATE \textbf{continue} 
    \ENDIF
    \STATE $D_s \leftarrow \{d \in S \mid d \notin \mathcal{E} \land d \neq s \land \text{CanFit}(d, s, P) \}$ \label{line:find_valid}
    \IF{$D_s \neq \emptyset$}
        \STATE $d^* \leftarrow \text{argmax}_{d \in D_s} \left( d.C_\text{used} + \sum_{(s',d') \in P, d'=d} s'.C_\text{used} \right)$ \label{line:best_fit}
        \STATE $P \leftarrow P \cup \{(s, d^*)\}$; $\mathcal{E} \leftarrow \mathcal{E} \cup \{s\}$
    \ENDIF
\ENDFOR
\RETURN $P$ \label{line:end_for}
\STATE {\bfseries Procedure} CanFit($d, s, P$): \label{line:can_start}
\STATE $\quad C_\text{load} \leftarrow d.C_\text{used} + \sum_{(s',d') \in P, d'=d} s'.C_\text{used}$
\STATE $\quad N_\text{load} \leftarrow d.N_\text{reqs} + \sum_{(s',d') \in P, d'=d} s'.N_\text{reqs}$
\STATE $\quad \textbf{return} \ (C_\text{load} + s.C_\text{used} \le C_\text{max}) \land (N_\text{load} + s.N_\text{reqs} \le B)$ \label{line:can_end}
\end{algorithmic}
\end{algorithm}

The core of the algorithm is an iterative matching process to match source rollouts to destinations (Lines~\ref{line:start_for}-\ref{line:end_for}).
For each source rollout, a set of valid destinations, $D_s$, are identified.
A valid destination rollout is a candidate rollout
which is not already slated for release and has sufficient headroom (Line~\ref{line:find_valid}), as verified by the \texttt{CanFit} procedure. %
In \texttt{CanFit}, the destination's current KVCache usage and request counts are summed with those from the source, plus any other workloads already assigned to this destination in the current plan $P$;
the projected total KVCache usage and request count
must not exceed %
 $C_\text{max}$ and $B$, respectively (Lines~\ref{line:can_start}-\ref{line:can_end}).
From this set of valid destinations, %
we select the one that will become most densely packed in terms of KVCache utilization after the trajectory transfers (Line~\ref{line:best_fit}).
This approach preserves capacity in other destinations for larger workloads.
If a valid destination is found, the trajectory move is recorded, and the source is marked for release.

The algorithm runs in $O(|S|^2)$ complexity, where $|S|$ is the number of candidate rollouts, efficient in practice due to the typically small size of $|S|$.
By creating larger batches on destination rollouts and maximizing the release of source rollouts, our algorithm simultaneously increases generation throughput and promotes the generation of on-policy data.

\vspace{-2mm}
\section{%
Trajectory-level Asynchrony Analysis} \label{sec:asynchrony_analysis}
Our asynchronous weight synchronization and repacking mechanism enable true trajectory-level asynchrony.
Trajectories are generated independently, each proceeding at its own pace with minimal bubbles in each rollout. 
This asynchrony gives rise to a key property of each trajectory's generation: \textit{inherent staleness}.
If a trajectory is generated using model version $K$
whose generation is finished when the actor model version becomes $M$, we define its inherent staleness as $M-K$.
This staleness is determined purely by the trajectory's generation latency and the trainer's model update speed.

\begin{figure}[t]
    \centering
    \vspace{-2mm}
    \includegraphics[width=\linewidth]{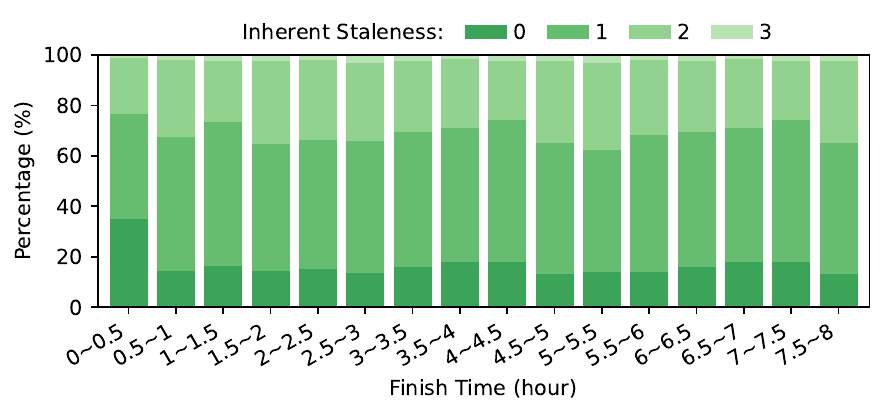}
    \vspace{-7mm}
    \caption{Inherent staleness distribution over finish time ranges of trajectory generation during RL training of a 7B model on 64 H800 GPUs using \sysname.
    }
    \label{fig:traj_dynamic}
    \vspace{-4mm}
\end{figure}

Under our fully decoupled/asynchronous design, \sysname{} does not require an explicitly configured staleness bound. Instead, the staleness %
of trajectories is naturally decided %
according to system dynamics %
(resource configuration and generation/training progress), eliminating any manual tuning of staleness bound. Especially, minimal inherent staleness is pursued with our rollout update design, as we eliminate idle time and update model version on each rollout as soon as its batch generation is completed or released. Figure~\ref{fig:traj_dynamic} shows that the inherent staleness of trajectories in \sysname{} remains consistently low (typically under 3).
\sysname{} fluidly caters to the shifting minimal staleness %
rendered by dynamic RL workloads, accommodating evolving trajectory lengths and variable generation latencies.

Upon completion, trajectories are moved to 
the experience buffer for trainer sampling.
Various sampling strategies can be adopted by the trainer (e.g., priority-based sampling~\cite{distrl, schaulPrioritizedExperienceReplay2016} on metrics like temporal-difference errors, importance ratio).
Experience sampling is a classic topic
in RL; %
developing effective sampling strategies utilizing massive experiences efficiently generated by \sysname{} is a 
promising direction for future exploration, and is orthogonal to the current paper.

%% file: tex/implementation.tex
\vspace{-2mm}
\section{Implementation} \label{sec:impl}
\sysname{} is implemented in \textasciitilde 11k lines of Python code (LoC). 

\noindent\textbf{Decopuled RL training framework.}
The trainer module and rollout module are implemented with 3.8k LoC on top of verl~\cite{hybridflow}, placed on different sets of GPUs.
The partial response pool stores the tokens and statistics of each ongoing trajectory from each rollout for fast recovery and online analysis.
Inter-module data transmission
is implemented with Ray~\cite{ray} through Remote Process Calls.

\noindent\textbf{Relay workers} are implemented in 1k LoC.
Each relay worker runs as a separate process, 
colocated with rollouts on the same GPU machine. 
For efficient intra-machine weight transfer, a relay worker uses pinned CPU shared memory.
This allows rollout workers to directly load model shards onto their GPUs via fast PCIe links.
We built a resilient communication layer 
for chain-based pipelined broadcast
using Unified Communication X (UCX)~\cite{ucx}. 
To ensure robust operation, the layer also integrates fault tolerance through heartbeat monitoring and dynamic chain rebuilding.

\noindent\textbf{Repack mechanism} is supported by the rollout manager, %
which is implemented with 1.6k LoC.

%% file: tex/experiment.tex
\vspace{-2mm}
\section{Evaluation} \label{sec:exp}
\textbf{Testbed.} We deploy \sysname{} on a cluster of 128 machines (1024 GPUs in total). Each machine is equipped with 8 NVIDIA H800-80GB GPUs inter-connected with 400GB/s NVLink. The inter-machine bandwidth is 8 $\times$ 400Gbps. Our experiments use the following software versions: CUDA 12.6, PyTorch 2.7.1, NCCL 2.26.2, and vLLM 0.9.0.

\noindent\textbf{Models.}
We choose Qwen2.5 models of sizes 7B, 32B, and 72B~\cite{qwen2.5}, which are popular models for RL post-training research in academia and industry.

\noindent\textbf{Baselines.}
We compare \sysname{} with four categories of RL post-training frameworks: \textit{synchronous}, \textit{one-step staleness}, \textit{stream generation}, 
and stream generation with \textit{partial rollout}:
\textbf{(1)} For the synchronous baseline, we use verl~\cite{hybridflow} v0.5.0, a state-of-the-art RL training framework, configured with its optimal colocate placement. 
\textbf{(2)} We implemented the asynchronous \textit{one-step staleness} and \textit{stream generation} pipelines ourselves on top of verl for fair comparison, alleviating implementation bias.
This is necessary as many prominent and concurrent asynchronous RL training systems are not open-sourced~\cite{streamrl, llamarl}, only runnable on Ascend NPUs~\cite{asyncflow}, or with limited system optimization~\cite{async-rlhf, rllm2025} in training, generation, and weight syncing stages. 
\textbf{(3)} For partial rollout, we use AReaL~\cite{areal} v0.3.0, a concurrent research that implements partial rollout with stream generation (truncating ongoing trajectory generation of rollouts and adopting updated weights to continue generation of these trajectories) and uses an algorithm
to mitigate the impact of different policy versions within each trajectory.

To ensure rigorous comparison, all baselines except AReaL
use vLLM~\cite{vllm} for generation and PyTorch FSDP~\cite{fsdp} for training. AReaL utilizes its modified SGLang~\cite{sglang} for generation
and only supports Megatron-LM~\cite{megatron-lm} for training.

\begin{figure*}[t]
    \hspace{-3mm}
    \subfigure[{\small 7B (1.12$\times$$\sim$5.48$\times$)}] {
        \includegraphics[width=0.32\linewidth]{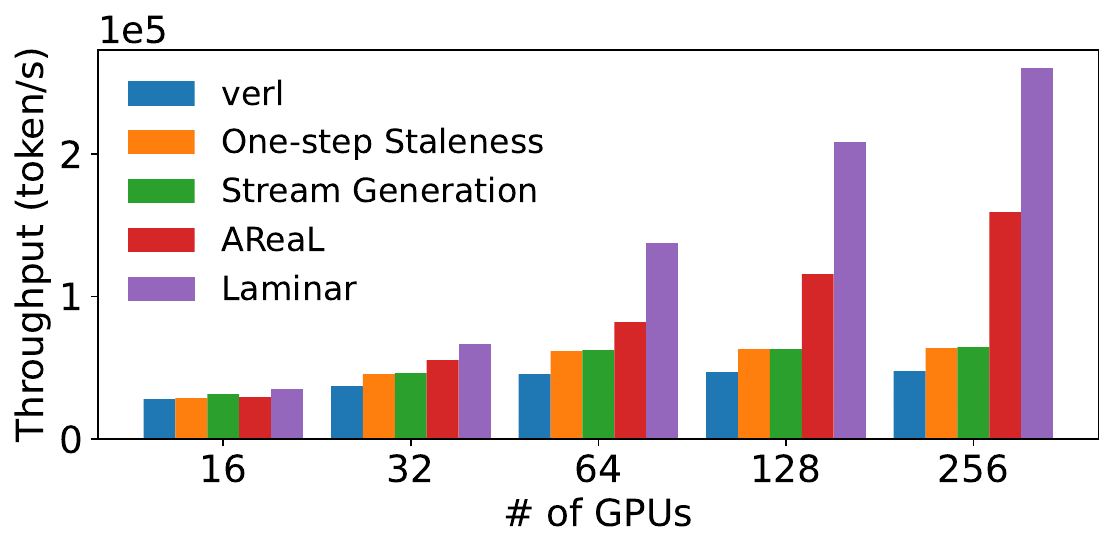}
    }
    \hspace{-3mm}
    \subfigure[{\small 32B (1.01$\times$$\sim$4.52$\times$)}] {
        \includegraphics[width=0.32\linewidth]{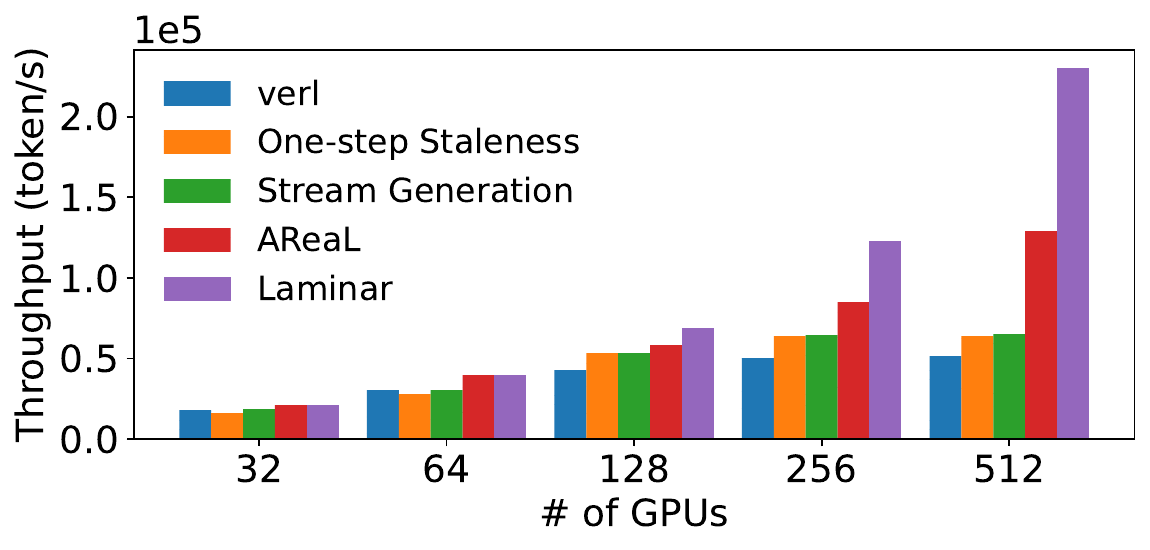}
    }
    \hspace{-3mm}
    \subfigure[{\small 72B (1.03$\times$$\sim$4.51$\times$)}] {
        \includegraphics[width=0.32\linewidth]{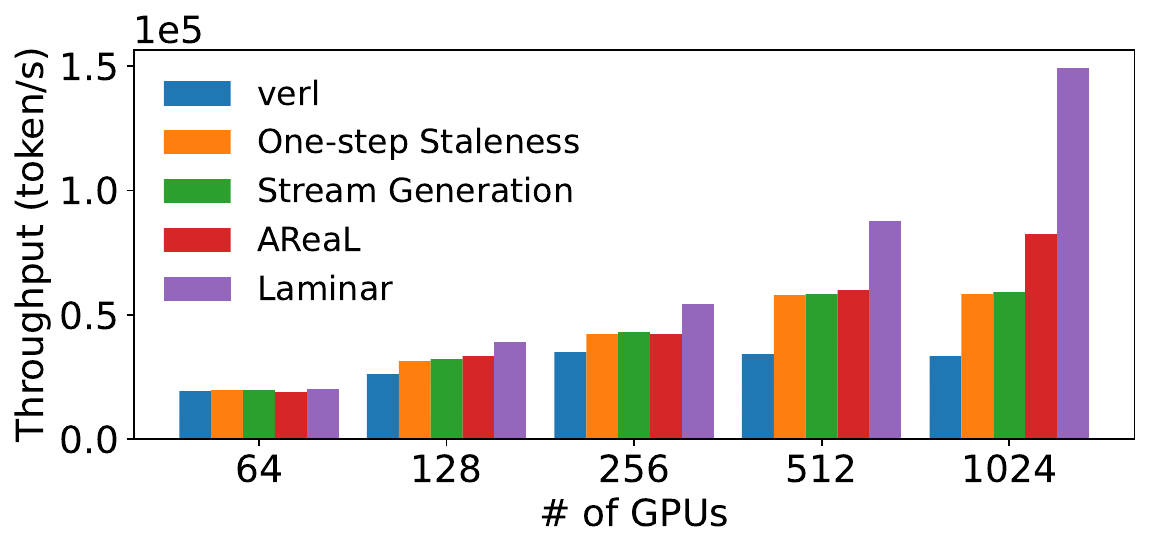}
    }
    \vspace{-3mm}
    \caption{Training throughput on single-turn task (mathematical reasoning).}
    \label{fig:throughput-math}
    \vspace{-1mm}
\end{figure*}

\begin{figure*}[t]
    \hspace{-3mm}
    \begin{minipage}{0.33\textwidth}
        \vspace{-1mm}
        \includegraphics[width=1\linewidth]{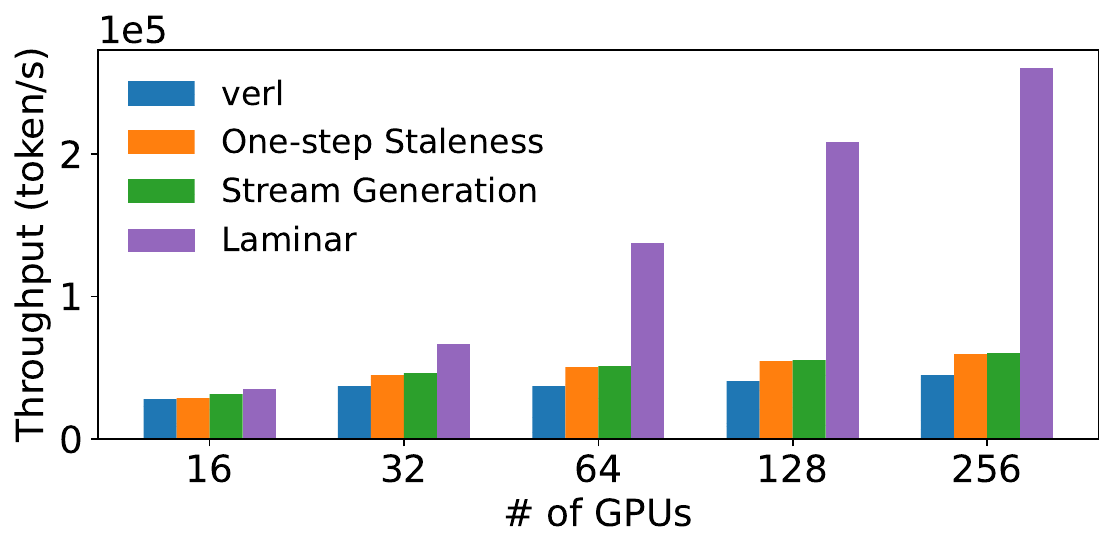}
        \vspace{-8mm}
        \caption{Training throughput on multi-turn task (tool calling). (7B, 1.21$\times$$\sim$5.42$\times$)}
        \label{fig:throughput-retool}
    \end{minipage}
    \hspace{-2mm}
    \begin{minipage}{0.65\textwidth}
        \vspace{-1.5mm}
        \subfigure[{\small 7B (32 GPUs)}] {
            \includegraphics[width=0.48\linewidth]{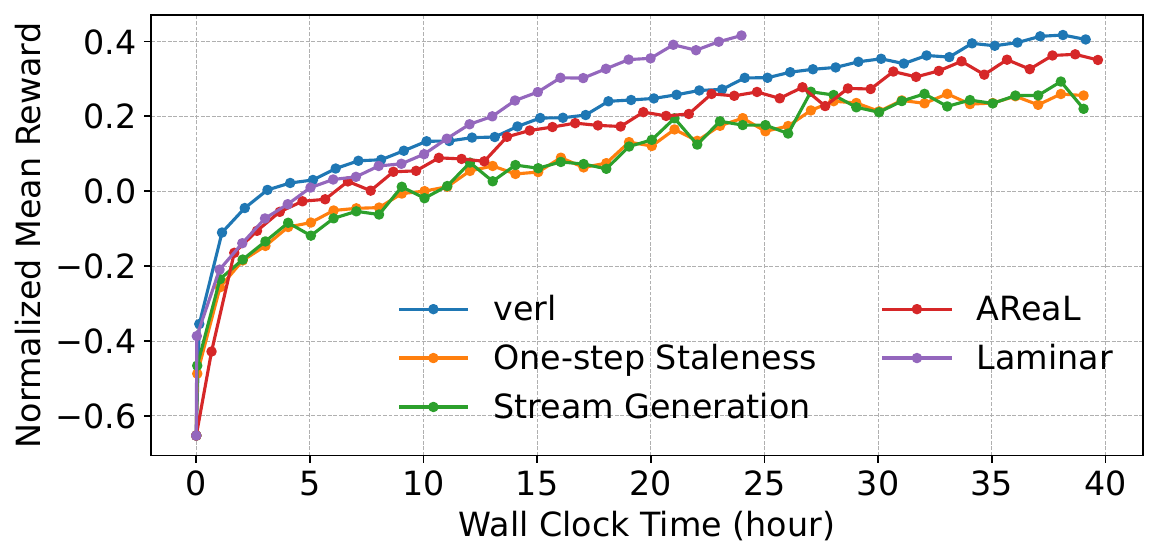}
        }
        \hspace{-3mm}
        \subfigure[{\small 32B (128 GPUs)}] {
            \includegraphics[width=0.48\linewidth]{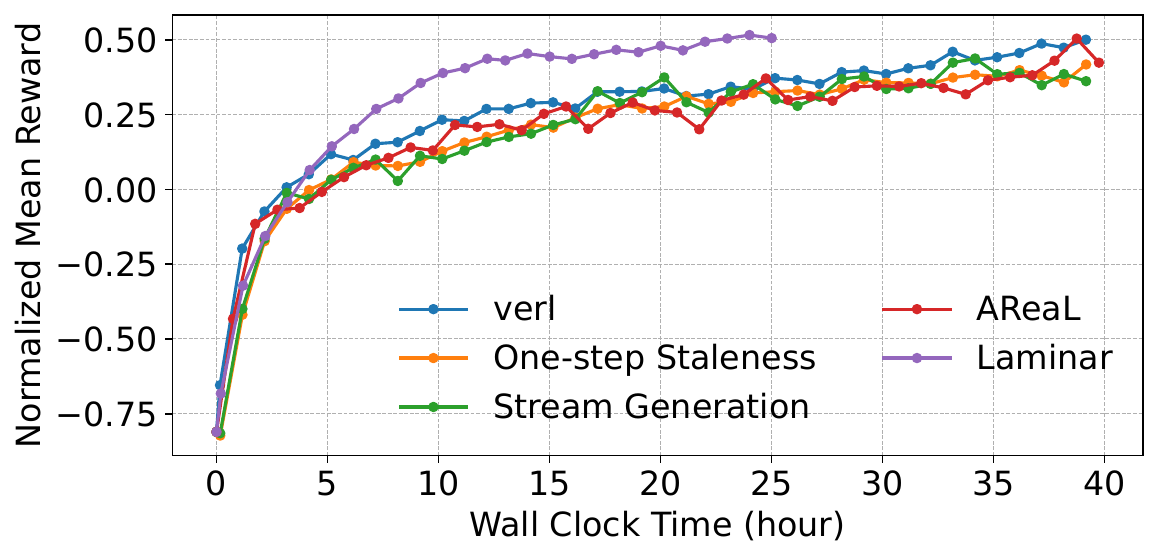}
        }
        \vspace{-5mm}
        \caption{Reward with respect to wall clock time.}
        \label{fig:eval_time_to_reward}
    \end{minipage}
    \vspace{-3mm}
\end{figure*}

\noindent\textbf{Datasets.}
We perform RL post-training on "DAPO-Math-17k" dataset~\cite{dapo}, which is widely used for math solving and multi-turn tool-calling tasks~\cite{retool,tool-survey}.
Maximum input and output lengths are set to 2K and 16K, respectively.
Distributions of model output length 
are given in 
Appendix~\ref{appendix:exp_response_dist}.

\noindent\textbf{Settings.}
We evaluate \sysname{} on both math reasoning and multi-turn tool-calling tasks.
For both tasks, we utilize an open-source implementation of GRPO algorithm~\cite{grpo} with a higher clipping range in the RL loss (i.e., Clip-Higher)~\cite{dapo}, which is widely adopted for RL post-training.
The effectiveness of \sysname{} does not rely on any specific RL algorithm and can generalize to others such as PPO~\cite{ppo}.
We utilize a rule-based reward function to 
score trajectories
on both tasks following~\cite{deepseek-r1, dapo, retool}.
For tool-calling task, the rollout interacts with a code sandbox~\cite{sandboxfusion} to generate reasoning steps
with multi-turn code execution for solving math problems.
The global training batch size is set to 8192 with 512 prompts each having 16 responses, and the number of mini-batch update steps per training iteration is 16; the maximum number of tool calls is set to 8, following previous research~\cite{dapo, retool}. 

\noindent\textbf{Metrics.}
We use throughput (tokens/sec) as the main performance metric~\cite{hybridflow}, computed by dividing the total number of tokens in prompts and responses in a global training batch by one RL iteration time (the duration between consecutive actor update completions).
All reported performance numbers are averaged over 5 RL iterations after a 10-iteration warm-up.

\vspace{-3mm}
\subsection{End-to-End Training Throughput}%
\label{sec:exp_e2e_train_performance}
Figures~\ref{fig:throughput-math}, \ref{fig:throughput-retool} show the training throughput of various RL frameworks under different model sizes.
We tune the optimal placement under each baseline by matching the generation and training throughput to reduce GPU idleness 
between the two stages.
The placement configurations are listed in Appendix~\ref{appendix:exp_hyperparam}, denoting a key performance factor when actor and rollouts are disaggregated~\cite{streamrl}.
We set the staleness bound of stream generation to 1 following~\cite{streamrl, asyncflow} and set
AReaL's to $\infty$ to achieve the largest training throughput.
The maximum staleness 
in \sysname{}, as we observed in all experiment settings, is 4. %
The global training batch size remains fixed when the cluster scales, a strong scaling setting which is standard practice in RL research~\cite{hybridflow, areal, realHF}.

\noindent\textbf{Overall performance.}
We observe that \sysname{} consistently outperforms the baselines across all model and cluster scales.
In Figure~\ref{fig:throughput-math}, \sysname{} achieves an average speedup of 2.56$\times$ (up to 5.49$\times$) over verl, 1.98$\times$ (up to 4.09$\times$) over one-step staleness pipeline, 1.93$\times$ (up to 4.06$\times$) over stream generation, and 1.39$\times$ (up to 1.81$\times$) over AReaL.
Similar gains are achieved on the tool calling task (Figure~\ref{fig:throughput-retool}), where \sysname{} delivers an average speedup of 2.62$\times$ across all baselines.
The performance advantage of \sysname{} increases as we scale to more GPUs, %
stemming from two primary factors.

{\em First}, \sysname{} adopts trajectory-level asynchrony, allowing each rollout to proceed at its own pace and enhancing generation throughput.
In contrast, baselines are constrained by a global synchronization barrier, which forces faster rollouts to wait for the slowest ones, creating significant pipeline bubbles. 
While AReaL attempts to mitigate this with partial rollouts, %
it introduces overhead due to trajectory generation aborting/resumption
and KVCache recomputation, which adversely affects generation throughput. %
{\em Second}, by minimizing rollout bubbles, \sysname{} sustains a higher generation throughput. This efficiency allows %
allocating more
GPUs to the trainer 
within a given cluster
as compared to other asynchronous baselines, while still balancing 
generation and training throughput.

\noindent\textbf{Scalability.}
Figures~\ref{fig:throughput-math}, \ref{fig:throughput-retool} show that \sysname{} achieves better scalability than all baselines,
a strong scaling efficiency
of 53.7\% (up to 68.2\% on 32B model) on math tasks and 46.5\% on tool-calling tasks. The scaling efficciency is computed by dividing $\frac{\mbox{throughput in largest scale}}{\mbox{throughput in smallest scale}}$ by $\frac{\mbox{max. \# of GPUs}}{\mbox{min. \# of GPUs}}$~\cite{amdahl1967strongscaling}.
The best baselines only reach 33.6\% (up to 39.2\%) (AReaL on math) and 12.9\% (stream generation on tool-calling).
The enhanced scalability of \sysname{} stems from its efficient resource utilization across all scales. 
As the cluster scales up, \sysname{} effectively utilizes additional GPUs to boost training throughput with little bubbles, as shown in Figure~\ref{fig:baselines}(e).
By avoiding straggler effects from global synchronization, \sysname{} exhibits increasing speedups as training scales. This results in an average speedup of 3.34$\times$ at the largest cluster scales across all baselines and model sizes. 
A more detailed analysis of this scalability is in Appendix~\ref{appendix:detailed_exp_analysis}.

\vspace{-2mm}
\subsection{%
Training Convergence} \label{sec:exp_train_convergence}
Model convergence speed is crucial in RL post-training, measured by the training reward improvement over time.
We compare \sysname{} with baselines using their respective tuned or open-sourced hyperparameters for the GRPO algorithm, with staleness bound under 1 for the stream generation baseline.
AReaL utilizes its proposed Decoupled PPO~\cite{hiltonBatchSizeinvariancePolicy2022} algorithm to mitigate errors introduced by partial rollout with a suggested optimal staleness of 4.
\sysname's maximum staleness is 4 for a direct comparison 
(detailed settings in Appendix~\ref{appendix:exp_hyperparam}).
As shown in Figure~\ref{fig:eval_time_to_reward}, \sysname{} converges about 1.77$\times$ faster for 7B and 1.59$\times$ faster for 32B than the best baseline (on-policy verl) on the math reasoning task.
\sysname{} achieves this by significantly increasing the training throughput while minimizing staleness without introducing additional biases; however, in other asynchronous systems, the throughput improvement is outweighed by staleness or additional biases, such as mixing multiple policy versions within a single trajectory in AReaL.

\begin{figure}[t]
    \subfigure[{\small 32B}] {
        \includegraphics[width=0.48\linewidth]{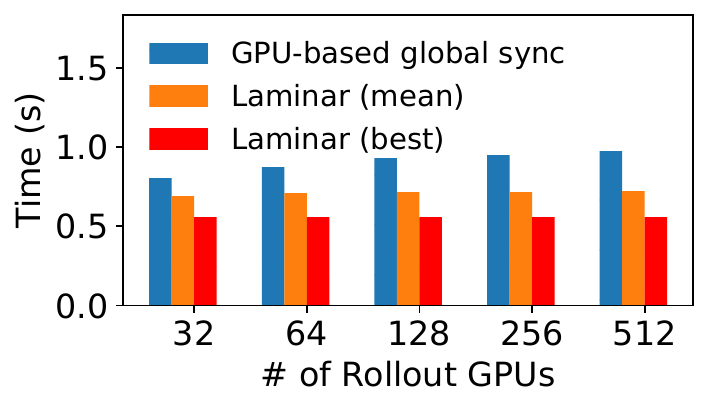}
    }
    \hspace{-3mm}
    \subfigure[{\small 72B}] {
        \includegraphics[width=0.48\linewidth]{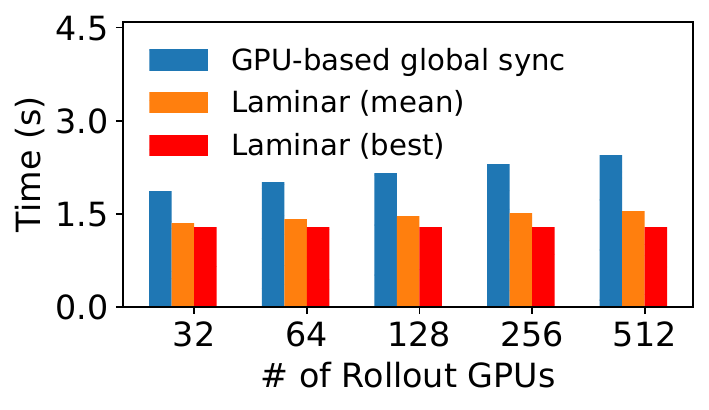}
    }
    \vspace{-5mm}
    \caption{Rollout waiting time during weight syncing. Trainer has the same number of GPUs as all rollouts' GPUs.} 
    \label{fig:exp_rollout_waiting_time}
    \vspace{-3mm}
\end{figure}

\begin{figure}[t]
    \centering
    \includegraphics[width=\linewidth]{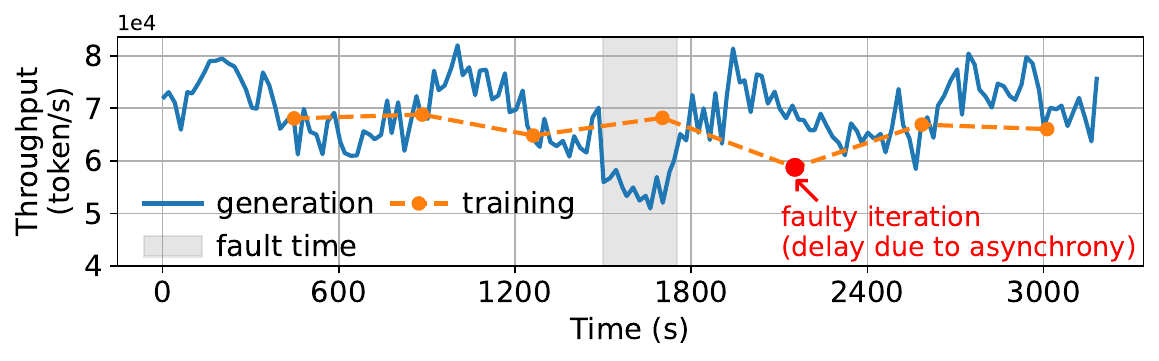}
    \vspace{-7mm}
    \caption{Training with a rollout machine failure.}
    \label{fig:exp_fault_tolerance_gen}
    \vspace{-3mm}
\end{figure}

\vspace{-2mm}
\subsection{%
Weight Synchronization Overhead} \label{sec:exp_weight_sync}

We evaluate the weight synchronization overhead of our relay worker design by measuring rollout waiting time and actor stalling time.
We compare \sysname{} against the GPU-based global synchronization approach adopted by recent RL systems~\cite{hybridflow, openrlhf, llamarl, streamrl, asyncflow}, 
that each actor model partition is broadcast to the corresponding model shards on each rollout. It follows the mapping between actor and rollout parallel groups and uses NCCL to achieve zero-copy transfer~\cite{llamarl, openrlhf}.

Figure~\ref{fig:exp_rollout_waiting_time} compares the rollout waiting time, measured 
from when each rollout begins updating to the latest weights until the update is complete.
From 64 to 1024 GPUs, \sysname{} consistently outperforms the GPU-based global synchronization.
It reduces the average and best-case waiting times by up to 37\% and 47\%, respectively. 
\sysname{}'s best-case occurs when the latest weights are already cached on the relay worker's CPU memory, allowing for loading model shards in parallel over the fast PCIe bus.
Notably, \sysname{}'s average rollout waiting time remains close to its best-case, %
due to our RDMA-based relay broadcast (Figure~\ref{fig:relay_broadcast_latency} in Appendix~\ref{sec:appendix_broadcast}) and trajectory-level asynchrony. 
By eliminating global synchronization points, few rollouts request the latest weights exactly when the actor finishes its update.
Consequently, most rollouts can fetch weights locally over PCIe
without waiting for a network broadcast.

\sysname{} also minimizes actor stalling time thanks to our hierarchical relay worker design.
The actor only transfers weights to a single master relay, and its communication overhead remains constant regardless of the number of rollout replicas. 
Therefore, the actor stalls only 0.64 and 1.40 seconds for 32B and 72B models, respectively.

\vspace{-2mm}
\subsection{Repack %
Efficiency} \label{sec:exp_repack}
We validate the benefit of our repack mechanism in eliminating remaining pipeline bubbles during generation. In this experiment, we use the same placement setting as the end-to-end experiment on the 32B model with 128 GPUs
(\textsection\ref{sec:exp_e2e_train_performance}): 
64 GPUs for the trainer and 64 GPUs for rollouts, with each rollout occupying 4 GPUs (totaling 16 rollouts).

\begin{figure}[t]
    \centering
    \includegraphics[width=\linewidth]{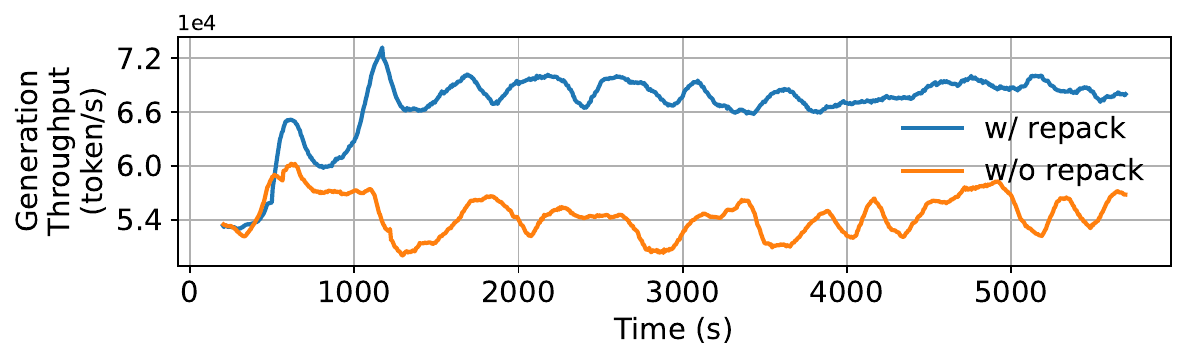}
    \vspace{-7mm}
    \caption{Repack efficiency.}
    \label{fig:exp_repack_throughput}
    \vspace{-3mm}
\end{figure}

As shown in Figure~\ref{fig:exp_repack_throughput}, enabling the repack mechanism boosts generation throughput by 26\% than without.
\sysname{} detects underutilized rollouts by monitoring their GPU KVCache utilization. %
As shown in Table~\ref{tab:repack_staistics}, the repack mechanism increases average KVCache utilization by 14.8\%. This improvement is achieved with a negligible repacking overhead of 0.69s and, crucially, does not increase the latency of trajectories generation.

\begin{table}
\caption{
Statistics of the rollouts with and without repack.
}
\vspace{-4mm}
\resizebox{\linewidth}{!}{
\begin{tabular}{c|c|c|c}
\toprule
\sysname{} & \makecell{Average/max latency of\\generating trajectories (s)} & \makecell{Repack\\overhead (s)} & \makecell{Average \\ KVCache util.} \\ 
\midrule
\makecell{w/ repack} & 290/828 & 0.69 & 82.2\% \\ 
\midrule
\makecell{w/o repack}   & 296/826   & /  & 71.6\% \\ 
\bottomrule                
\end{tabular}
}
\label{tab:repack_staistics}
\vspace{-3mm}
\end{table}

\vspace{-2mm}
\subsection{Fault Tolerance Analysis} \label{sec:exp_fault_tolerance}

We evaluate the robustness of \sysname{} when encountering failures during RL training. 
We manually kill a rollout machine containing two replicas during a training job that has the same setting as in \textsection\ref{sec:exp_repack}.
As shown in Figure~\ref{fig:exp_fault_tolerance_gen}, this event causes an immediate drop in generation throughput due to the loss of rollouts.
Training throughput remains unaffected or experiences a slight drop while waiting for adequate trajectories from remaining healthy rollouts.
The system recovers in approximately 252 seconds, including allocating a new machine and initializing the rollouts on it.
After recovery, both generation throughput and training throughput recover to original levels.
This demonstrates that \sysname{} handles machine failures without disrupting training progress.

%% file: appendix/appendix_related_work.tex
\section{Related Work}\label{appendix:related_work}

\noindent \textbf{RL frameworks for LLMs.} 
Early reinforcement learning (RL) frameworks for LLMs primarily focused on synchronous algorithms~\cite{CollosalChat,vonwerra2022trl,deepspeed-chat,nemo-aligner,adaptive-rlhf,openrlhf,hybridflow,realHF,k1-5,nemo-rl,rlhfuse,griggs2025skrylv01,slime_github,deepseek-r1}. More recent work has begun to explore asynchronous RL~\cite{distrl,rllm2025,streamrl,intellect-2,llamarl,asyncflow}. 
However, these emerging asynchronous systems still depend on a global weight synchronization point. This design forces faster rollouts to idle while waiting for stragglers, creating significant long-tail performance bubbles. 
Some systems attempt to mitigate this issue with partial rollouts~\cite{k1-5, seed1.5,llamarl,areal}.
But this technique introduces severe penalties: the high overhead of recomputing KVCache after each update and the undesirable mixing of multiple policy versions within a single trajectory. These issues are notably exacerbated at scale. In contrast, \sysname{}'s fully decoupled architecture and trajectory-level asynchrony eliminates these long-tail bubbles while guaranteeing that each trajectory is generated with a single, consistent policy version.

\noindent \textbf{Asynchronous deep RL systems.} Extensive research exists on asynchronous deep RL systems~\cite{openaibaseline,hafner2017tensorflowagents,torchrl,espeholtIMPALAScalableDistributed2018,liang2018rllib,liang2021rllibflow,weng2022tianshou,alphastar,dota2}. However, they target small-scale DNNs, not designed for the efficient inference, generation, and training patterns of modern LLMs.
Some systems employ fully asynchronous parameter updates where gradients are computed from stale parameters~\cite{benjamin2011hogwild,mnih2015dqn,Mnih:2016qfq}.
This approach can introduce training instability and is thus rarely used for LLMs.
The success of state-of-the-art deep RL systems like AlphaStar~\cite{alphastar} and OpenAI Five~\cite{openaifive} has been attributed to massive scale-out, with OpenAI Five, for instance, leveraging 57,200 rollout workers on 51,200 CPUs for 10 months~\cite{dota2}. This emphasis on scale has also spurred optimizations in related components, such as distributed experience buffers~\cite{horganDistributedPrioritizedExperience2018,cassirer2021reverb,wangGEARGPUCentricExperience2023}.

\noindent \textbf{Fault-tolerant training.}
Previous studies~\cite{robust-tpu, robust-shlab, megascale, aegis,mycroft,byterobust} have introduced robust AI infrastructures to automatically detect and localize various training failures through real-time monitoring and stop-time diagnosis.
The sophisticated techniques they rely on, such as RDMA-level metric collection and \texttt{dmesg} inspection, can be paired with our heartbeat-based failover mechanism to improve observability of rollout replicas and trainer workers in the RL system further.
Many diagnostic tools have been developed to pinpoint network faults, both inter-host~\cite{everflow,lossradar,netbouncer} and intra-host~\cite{hostping}.
Rather than delving deep into the network stack to expose the exact root cause, our broadcast chain rebuilding mechanism works around communication failures by routing around faulty relay workers during weight synchronization.
Checkpointing optimizations~\cite{checkfreq,check-n-run,gemini-aws,bytecheckpoint} further mitigate stalls and support flexible transfer~\cite{bytecheckpoint, wagenlander2024tenplex}.
By incorporating these methods, our trainer can recover from failures more quickly.

%% file: tex/conclusion.tex
\vspace{-2mm}
\section{Conclusion} \label{sec:conclusion}

\sysname{} is an RL framework that addresses the limited scalability and long-tail trajectory generation problem in LLM post-training.
We enable trajectory-level asynchrony, with each trajectory generated and consumed independently at its own optimal pace, eliminating rigid global synchronization.
This design is achieved through a fully decoupled architecture with relay workers enabling asynchronous weight synchronization, allowing rollouts to pull new model parameters anytime without stalling computation.
This architecture accommodates evolving trajectory lengths in RL training, while isolating component failures to ensure robust fault tolerance for long-running jobs.
Our dynamic repack mechanism further consolidates long-tail trajectories to maximize generation throughput while minimizing staleness.
We conducted extensive evaluation at scales up to 1024 GPUs. Our results demonstrate that \sysname{} achieves up to 5.48$\times$ speedup over state-of-the-art systems and reduces model convergence time.

%% file: appendix/appendix_experiement_detail.tex
\section{Experiment Details} \label{appendix:exp_detail}

\subsection{Response Length Distribution of Each Model}
\label{appendix:exp_response_dist}

\begin{figure}[htbp]
\centering
\subfigure[Model: 7B, Task: Math]{
    \includegraphics[width=0.2\textwidth]{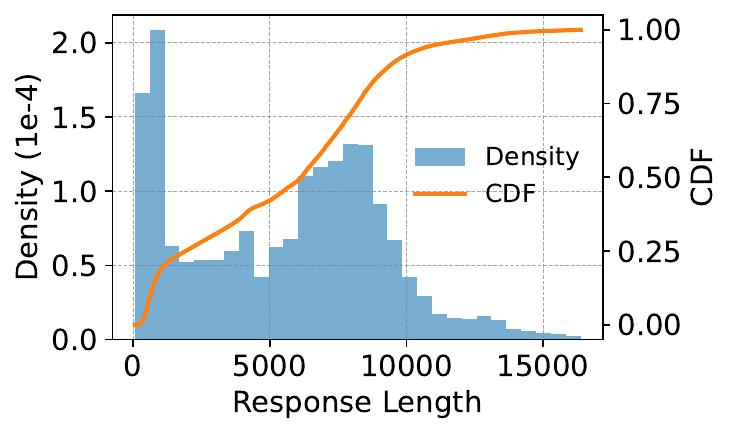}
}
\subfigure[Model: 32B, Task: Math]{
    \includegraphics[width=0.2\textwidth]{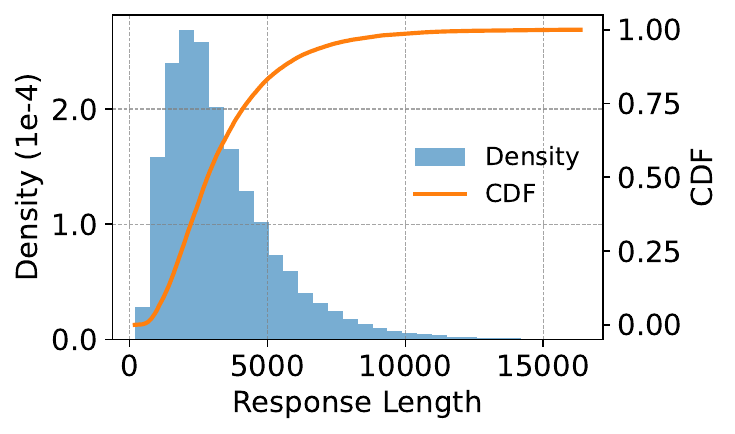}
}
\subfigure[Model: 72B, Task: Math]{
    \includegraphics[width=0.2\textwidth]{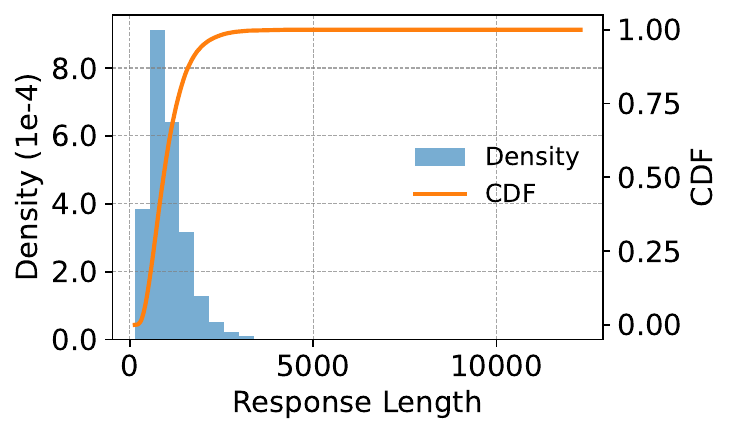}
}
\subfigure[Model: 7B, Task: Tool]{
    \includegraphics[width=0.2\textwidth]{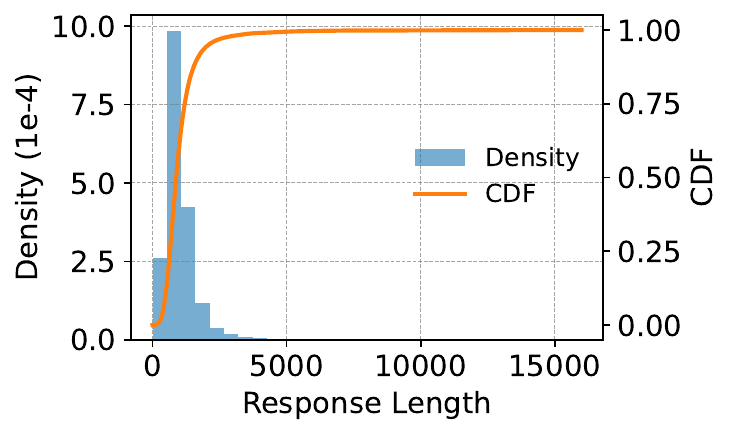}
}

\caption{The response length distributions on the DAPO-Math-17k dataset of the model checkpoints used in the throughput experiments. 
}
\label{fig:throughput_outlen_distribution}
\end{figure}

The response length distributions of each model used in the throughput experiments are listed in Figure~\ref{fig:throughput_outlen_distribution}.

\subsection{Hyperparameter Details}
\label{appendix:exp_hyperparam}

\begin{table*}[htbp]
\caption{GPU allocation of the systems in the throughput experiments. For verl, we adopt the colocated allocation which uses all the GPUs for train and rollout alternately.}
\label{tab:allocation}
\centering
\begin{tabular}{crrrrrrrrr}
\toprule
\multirow{2}{*}{\textbf{System}} & \multicolumn{3}{c}{\textbf{7B}} & \multicolumn{3}{c}{\textbf{32B}} & \multicolumn{3}{c}{\textbf{72B}} \\
\cmidrule(lr){2-4} \cmidrule(lr){5-7} \cmidrule(lr){8-10}
 & \textbf{Total} & \textbf{Train} & \textbf{Rollout} & \textbf{Total} & \textbf{Train} & \textbf{Rollout} & \textbf{Total} & \textbf{Train} & \textbf{Rollout} \\
\midrule
\multirow{5}{*}{\makecell{verl}} & 16 & \multicolumn{2}{c}{\multirow{5}{*}{Colocated}} & 32 & \multicolumn{2}{c}{\multirow{5}{*}{Colocated}} & 64 & \multicolumn{2}{c}{\multirow{5}{*}{Colocated}} \\
& 32 & & & 64 & & & 128 & & \\
& 64 & & & 128 & & & 256 & & \\
& 128 & & & 256 & & & 512 & & \\
& 256 & & & 512 & & & 1024 & & \\
\midrule
\multirow{5}{*}{\makecell{One-step \\ Staleness}} & 16 & 8 & 8 & 32 & 16 & 16 & 64 & 32 & 32 \\
& 32 & 8 & 24 & 64 & 32 & 32 & 128 & 64 & 64 \\
& 64 & 16 & 48 & 128 & 48 & 80 & 256 & 96 & 160 \\
& 128 & 32 & 96 & 256 & 64 & 192 & 512 & 192 & 320 \\
& 256 & 40 & 216 & 512 & 80 & 432 & 1024 & 256 & 768 \\
\midrule
\multirow{5}{*}{\makecell{Stream \\ Generation}} & 16 & 8 & 8 & 32 & 16 & 16 & 64 & 32 & 32 \\
& 32 & 8 & 24 & 64 & 32 & 32 & 128 & 64 & 64 \\
& 64 & 16 & 48 & 128 & 48 & 80 & 256 & 96 & 160 \\
& 128 & 32 & 96 & 256 & 64 & 192 & 512 & 192 & 320 \\
& 256 & 40 & 216 & 512 & 80 & 432 & 1024 & 256 & 768 \\
\midrule
\multirow{5}{*}{\makecell{AReaL}} & 16 & 8 & 8 & 32 & 16 & 16 & 64 & 32 & 32 \\
& 32 & 16 & 16 & 64 & 32 & 32 & 128 & 64 & 64 \\
& 64 & 32 & 32 & 128 & 64 & 64 & 256 & 128 & 128 \\
& 128 & 64 & 64 & 256 & 128 & 128 & 512 & 320 & 192 \\
& 256 & 128 & 128 & 512 & 256 & 256 & 1024 & 640 & 384 \\
\midrule
\multirow{5}{*}{\makecell{\sysname}} & 16 & 8 & 8 & 32 & 16 & 16 & 64 & 32 & 32 \\
& 32 & 24 & 8 & 64 & 32 & 32 & 128 & 64 & 64 \\
& 64 & 40 & 24 & 128 & 64 & 64 & 256 & 128 & 128 \\
& 128 & 80 & 48 & 256 & 128 & 128 & 512 & 320 & 192 \\
& 256 & 192 & 64 & 512 & 256 & 256 & 1024 & 768 & 256 \\
\bottomrule
\end{tabular}
\end{table*}

\noindent\textbf{Throughput experiments.}
We evaluate system throughput on two tasks: math and tool-calling. For the math task, we use three intermediate checkpoints of different scales during the RL training from \texttt{Qwen2.5-Math-7B}, \texttt{Qwen2.5-32B}, and \texttt{Qwen2.5-Math-72B}, respectively. For the tool-calling task, we use the 7B checkpoint for the RL training in the open-source implementation of ReTool. Table~\ref{tab:allocation} details the GPU allocation for each experiment, which we tuned to maximize throughput in our environment.

During rollout, we use temperature sampling with $t=1$. We do not use top-$P$ or top-$k$ sampling. For verl, One-step Staleness, and Stream Generation, the global rollout batch size is 8192, matching the training batch size. For AReaL and \sysname{}, we set the maximum concurrency to 1024 trajectories per rollout replica.
The rollout tensor parallelism (TP) sizes are set to 4 and 8 for the 32B and 72B models, respectively. For the 7B model, the size varies according to the system's throughput characteristics. In AReaL and \sysname{}, we set the TP size to 1 to maximize throughput. In verl, One-step Staleness, and Stream Generation, we set the TP size to 2, which balances generation throughput against latency, mitigating the long-tail bottlenecks.

For AReaL, we use Megatron-LM's hybrid of data (DP), tensor (TP), and pipeline parallelism (PP). The TP and PP sizes are set to 2 and 1 for the 7B model, 4 and 2 for the 32B model, and 4 and 4 for the 72B model. The DP size is adaptively set to $\frac{\text{\# of Train GPUs}}{\text{TP Size} \times \text{PP Size}}$. For other systems, we combine Torch DDP, FSDP, and Ulysses Sequence Parallelism (SP). The FSDP and SP sizes are set to 8 and 4 for the 7B model, 16 and 8 for the 32B model, and 32 and 8 for the 72B model. The DDP size is set to $\frac{\text{\# of Train GPUs}}{\text{FSDP Size}}$.

\noindent\textbf{Convergence experiments.}
We conduct convergence experiments using two base models: \texttt{Qwen2.5-Math-7B} and \texttt{Qwen2.5-32B}. The GPU allocation is identical to that of the throughput experiments. Table~\ref{tab:convergence_exp} lists the other hyperparameters. For AReaL and \sysname{}, we reduce the maximum per-rollout concurrency to 256. 
We also adopt the default FIFO sampling strategy for these two systems.

\begin{table*}[htbp]
\centering
\caption{Hyperparameters for convergence experiments. Configurations are based on the original AReaL paper and DAPO. For asynchronous systems (One-step Staleness, Stream Generation, and \sysname), we increase the mini-batch size to 2048 to stabilize training with off-policy data, aligning with AReaL's configuration.}
\label{tab:convergence_exp}
\begin{tabular}{lccccc}
\toprule
\textbf{System} & \textbf{verl} & \makecell{\textbf{One-step}\\\textbf{Staleness}} & \makecell{\textbf{Stream}\\\textbf{Generation}} & \textbf{AReaL} & \textbf{\sysname} \\
\midrule
\multirow{2}{*}{Algorithm} & \multirow{2}{*}{GRPO} & \multirow{2}{*}{GRPO} & \multirow{2}{*}{GRPO} & Decoupled & \multirow{2}{*}{GRPO} \\
& & & & PPO & \\
Learning Rate & 1e-6 & 1e-6 & 1e-6 & 2e-5 & 1e-6 \\
Weight Decay & 0.1 & 0.1 & 0.1 & 0.05 & 0.1 \\
Clip $\varepsilon_{\text{high}}$ & 0.28 & 0.28 & 0.28 & 0.2 & 0.28 \\
Clip $\varepsilon_{\text{low}}$ & 0.2 & 0.2 & 0.2 & 0.2 & 0.2 \\
Discount $\gamma$ & 1.0 & 1.0 & 1.0 & 1.0 & 1.0 \\
GAE $\lambda$ & 1.0 & 1.0 & 1.0 & 1.0 & 1.0 \\
Group Size & 16 & 16 & 16 & 16 & 16 \\
Training Global Batch Size & 8192 & 8192 & 8192 & 8192 & 8192 \\
Training Mini-Batch Size & 512 & 2048 & 2048 & 2048 & 2048 \\
Per-rollout Max Concurrency & N/A & N/A & N/A & 256 & 256 \\
Sampling Strategy & N/A & N/A & N/A & FIFO & FIFO \\
Max Staleness Bound & 0 & 1 & 1 & 4 & 4 (observed) \\
\bottomrule
\end{tabular}
\end{table*}

%% file: appendix/appendix_experiment_analysis.tex
\section{Detailed Experiment Analysis} \label{appendix:detailed_exp_analysis}

\noindent\textbf{Performance in small cluster scales.}
In relatively small cluster scales, \sysname{} provides a moderate but consistent 1.41$\times$ speedup on average across all the baselines and model sizes on at most 64 GPUs, as shown in Figure~\ref{fig:throughput-math}.
The performance in these configurations is constrained by the fact that both training and rollout of LLMs require a number of GPUs to accommodate the model via parallelism. This severely limits the possible resource allocation choices, forcing different systems with similar placements, as shown in Table \ref{tab:allocation}. Furthermore, these placements are usually suboptimal since the training and rollout throughput can not be effectively balanced. Within such limited configurations, \sysname{} improves the throughput mainly by eliminating long-tail bubbles.

\noindent\textbf{Performance in large-scale clusters.}
The performance advantage of \sysname{} becomes much more pronounced in large-scale clusters, achieving an average speedup of 3.34$\times$ at the largest cluster scales across all the baselines and model sizes, as is shown in Figure~\ref{fig:throughput-math}.
Systems like verl, one-step staleness, and stream generation are heavily bottlenecked by the global weight synchronization. 
With the end-to-end latency of the generation stage bound by the long-tail trajectories, adding more rollout resources provides marginal returns on throughput.
AReaL mitigates this bottleneck by applying partial rollout, but its scalability is then limited by the KVCache recomputation overhead. As the cluster size scales up, this overhead worsens as the number of trajectories and the model update frequency increase, capping the system's performance.
In contrast, \sysname{} effectively resolves the global weight synchronization constraint and eliminates the long-tail bubbles while introducing minimal overhead through the relay design and repack mechanism. Furthermore, as the cluster size scales up, the repacking becomes increasingly effective with more rollout replicas to manage.

%% file: appendix/appendix_discussion.tex
\section{Discussion} \label{sec:appenix_discussion}
\noindent\textbf{Partial rollout.}
\sysname's trajectory-level asynchrony effectively mitigates long-tail latency bubbles by decoupling trajectory generation from global synchronization. 
Recently, an orthogonal approach has emerged that 
updates weights mid-generation to reduce these bubbles
known as partial rollout systems~\cite{k1-5, seed1.5, areal, llamarl}. 
These systems interrupt all ongoing generation and update rollout replicas once the latest actor weights become available.
The aborted trajectories then continue generating with the new weights. 
This technique can be integrated with synchronous, k-step staleness, and \sysname{} to further reduce idle time.
However, it introduces a challenge to training stability.
Updating weights amid generation
creates mixed-version trajectories, where a single long trajectory is produced using several successive weight versions. The expected number of versions grows linearly with sequence length and the actor-update frequency. These mixed-version trajectories distort the natural output-length distribution. They can also hinder convergence in valueless RL algorithms, which rely on forming trajectory groups from a consistent weight version. As update frequency rises, the contamination worsens, creating an undesirable coupling between system throughput and training stability. As shown in Figure~\ref{fig:eval_time_to_reward}, applying partial rollout can converge at a slower speed, 
requiring more algorithmic research to stabilize such training.

\noindent\textbf{Effective experiences sampling.}
In \sysname, as scaling out RL tasks, 
the scaling of trajectory generation allows rollout throughput to outpace the trainer's consumption speed (analyzed in \textsection\ref{sec:exp_e2e_train_performance}).
This creates an abundance of diverse experiences, but presents a key challenge: \textit{how to utilize these experiences effectively}. 
Experiences sampling is a classic problem
in traditional RL.
For instance, OpenAI's work on Dota 2 highlighted the negative impact of data staleness on training speed~\cite{dota2}.
Recent research advocates for priority-based sampling~\cite{distrl, espeholtIMPALAScalableDistributed2018, schaulPrioritizedExperienceReplay2016}, which considers metrics like TD errors, importance sampling, and entropy to 
prioritize the most informative transitions.

However,
experience sampling remains an underexplored area in large-scale LLM post-training. 
The ability of \sysname{} to efficiently generate massive experiences is a key strength, but this potential can only be fully realized with an effective sampling mechanism,
which is crucial for translating 
diverse experiences
and avoid learning on low-utility experiences.
Thus, developing effective sampling strategies for utilizing generated experiences represents a 
promising area for future work in large-scale RL systems.

%% file: appendix/appendix_comm.tex
\section{Theoretical Analysis of Chain-Based Pipelined Broadcast} \label{sec:appendix_broadcast}

We provide a formal analysis of the latency for a chain-based pipelined broadcast. The goal is to show that for large messages, such as LLM weights, the total broadcast time is dominated by a term independent of the number of nodes, making the approach highly scalable.

\vspace{-2mm}
\subsection{Communication Model}

We model the broadcast as a linear pipeline, where a master relay sends a message to $p-1$ other relays organized in a logical chain. Let the total number of nodes (master + relays) be $p$. The communication cost between any two adjacent nodes is defined by two parameters:
    
\noindent$\bullet$
$T_{\text{start}}$: The startup latency for sending a message, independent of its size. This includes costs like connection setup and initial processing.

\noindent$\bullet$
$T_{\text{byte}}$: The per-byte transmission time, determined by the network bandwidth ($T_{\text{byte}} = 1 / \text{Bandwidth}$).
The time $t$ to transmit a single message of size $s$ between two nodes is $t = s \cdot T_{\text{byte}} + T_{\text{start}}$.

In our pipelined approach, the total model weights, of size $M$ bytes, are divided into $k$ smaller chunks, each of size $M/k$.

\begin{figure}[t]
    \centering
    \includegraphics[width=\linewidth]{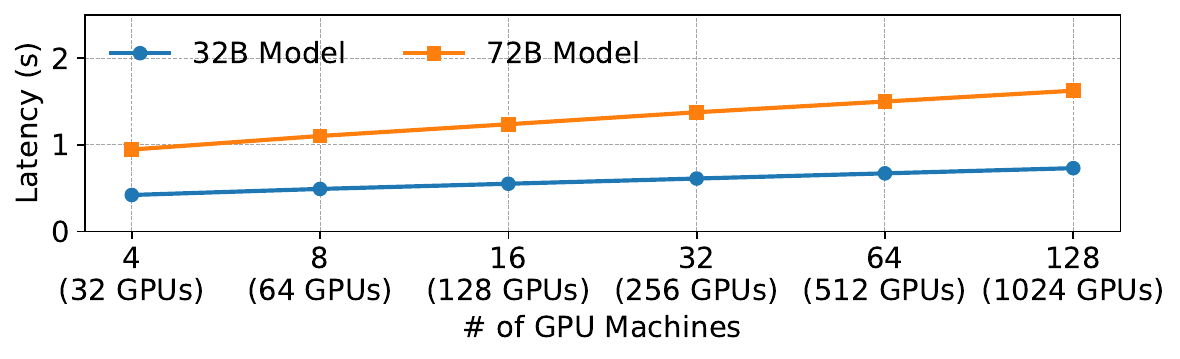}
    \vspace{-7mm}
    \caption{Relay broadcast latency with testbed in \textsection\ref{sec:exp}.}
    \label{fig:relay_broadcast_latency}
    \vspace{-3mm}
\end{figure}

\vspace{-2mm}
\subsection{Latency Derivation}

The total time for the broadcast, $T(p, k)$, is the time from when the master sends the first chunk until the last relay receives the last chunk. This can be modeled as the sum of two components:
\begin{enumerate}[leftmargin=0.5cm]
    \item 
The time for the first chunk to travel down the entire chain of $p-1$ hops.
    \item
The time for the remaining $k-1$ chunks to be received by the final relay after it has received the first one.
\end{enumerate}
The time to transmit one chunk between two nodes is $t_{\text{chunk}} = \frac{M}{k} T_{\text{byte}} + T_{\text{start}}$.
The first chunk must traverse $p-1$ network hops to reach the final relay. Due to the pipelined nature, this takes $(p-1) \times t_{\text{chunk}}$. Once the first chunk arrives at the final relay, the subsequent $k-1$ chunks arrive sequentially, with one chunk arriving every $t_{\text{chunk}}$. This adds $(k-1) \times t_{\text{chunk}}$ to the total time.
Therefore, the total latency is:
\begin{align*}
T(p, k) &= (p-1) \cdot t_{\text{chunk}} + (k-1) \cdot t_{\text{chunk}} \\
&= (p + k - 2) \cdot t_{\text{chunk}} \\
&= (p + k - 2) \left( \frac{M}{k} T_{\text{byte}} + T_{\text{start}} \right)
\end{align*}
Expanding this expression:
\begin{equation}
T(p, k) = M T_{\text{byte}} + (p-2)T_{\text{start}} + k T_{\text{start}} + \frac{(p-2)M}{k} T_{\text{byte}}
\label{eq:expanded}
\end{equation}

\vspace{-2mm}
\subsection{Scalability Analysis}

To understand the scalability with respect to the number of nodes $p$, we analyze the terms in Equation \ref{eq:expanded}. The choice of $k$ (the number of chunks) is critical. We can find the optimal $k$ that minimizes $T(p, k)$ by taking the derivative with respect to $k$ and setting it to zero:
\begin{align*}
\frac{\partial T}{\partial k} &= T_{\text{start}} - \frac{(p-2)M}{k^2} T_{\text{byte}} = 0 \\
k^2 &= \frac{(p-2)M T_{\text{byte}}}{T_{\text{start}}} \\
k^{*} &= \sqrt{\frac{(p-2)M T_{\text{byte}}}{T_{\text{start}}}}
\end{align*}
Substituting $k^{*}$ back into Equation \ref{eq:expanded}, we find the minimum possible broadcast time for a given $p$ and $M$:
\begin{align*}
T^{*}(p) &= M T_{\text{byte}} + (p-2)T_{\text{start}} + 2\sqrt{(p-2)M T_{\text{byte}} T_{\text{start}}}
\end{align*}
Let's analyze the composition of this optimal time:
\[ T^{*}(p) = \underbrace{M T_{\text{byte}}}_{\text{Bandwidth Term}} + \underbrace{(p-2) T_{\text{start}}}_{\text{Latency Term}} + \underbrace{2\sqrt{(p-2)M T_{\text{byte}} T_{\text{start}}}}_{\text{Pipeline Term}} \]

In the context of distributing large language models:

\noindent$\bullet$  
\textbf{The message size $M$ is very large} (e.g., 140 GB for a 70B model using BF16 precision).

\noindent$\bullet$
\textbf{The startup latency $T_{\text{start}}$ is very small} for RDMA (on the order of microseconds).
Consequently:
\begin{enumerate}[leftmargin=0.5cm]
    \item The \textbf{Bandwidth Term ($M T_{\text{byte}}$)} is the time to serialize the entire model onto a single network link. Given the large $M$, this term is the dominant component of the total latency and is completely independent of $p$.
    \item The \textbf{Latency Term ($(p-2) T_{\text{start}}$)} grows linearly with $p$. However, because $T_{\text{start}}$ is extremely small, this term's contribution is negligible even for thousands of nodes (e.g., $2000 \times 5\mu s = 10ms$).
    \item The \textbf{Pipeline Term} grows sub-linearly with the number of nodes ($O(\sqrt{p})$). While it grows faster than the latency term, its contribution remains significantly smaller than the bandwidth term for realistic system parameters.
\end{enumerate}

\noindent\textbf{Conclusion:} The total broadcast time is overwhelmingly dominated by the constant bandwidth term ($M T_{\text{byte}}$). The terms dependent on the number of nodes $p$ either have a very small coefficient ($T_{\text{start}}$) or grow sub-linearly ($O(\sqrt{p})$). This makes the total latency largely insensitive to the number of relays, proving that the chain-based pipelined broadcast is a highly scalable mechanism for distributing large model weights.